\documentclass[10pt,twocolumn,letterpaper]{article}

\usepackage{iccv}              

\definecolor{iccvblue}{rgb}{0.21,0.49,0.74}
\usepackage[pagebackref,breaklinks,colorlinks,allcolors=iccvblue]{hyperref}

\usepackage{multirow} 
\usepackage{colortbl}  
\usepackage{xcolor}
\usepackage{array}   
\usepackage{graphicx}
\usepackage{booktabs}
\usepackage{amsmath}

\def\paperID{2956} 
\def\confName{ICCV}
\def\confYear{2025}

\title{Dissecting Generalized Category Discovery:\\Multiplex Consensus under Self-Deconstruction}

\author{Luyao Tang$^{1,2,5*}$, Kunze Huang$^{1,2,*}$, Chaoqi Chen$^{3,\dag}$, Yuxuan Yuan$^{1}$, Chenxin Li$^{4}$\\Xiaotong Tu$^{1,2}$, Xinghao Ding$^{1,2}$  and Yue Huang$^{1,2}$\\
$^1$ Key Laboratory of Multimedia Trusted Perception and Efficient Computing, \\Ministry of Education of China, Xiamen University
{$^2$ School of Informatics, Xiamen University}\\
{$^3$ Shenzhen University}\quad
{$^4$ The Chinese University of Hong Kong}\quad
{$^5$ The University of Hong Kong}\\
{\tt\small  \{lytang, kzhuang\}@stu.xmu.edu.cn, cqchen1994@gmail.com,  yhuang2010@xmu.edu.cn}
}

\begin{document}
\maketitle
\begin{abstract}
Human perceptual systems excel at inducing and recognizing objects across both known and novel categories, a capability far beyond current machine learning frameworks. While generalized category discovery (GCD) aims to bridge this gap, existing methods predominantly focus on optimizing objective functions. We present an orthogonal solution, inspired by the human cognitive process for novel object understanding: decomposing objects into visual primitives and establishing cross-knowledge comparisons. We propose \textbf{ConGCD}, which establishes primitive-oriented representations through high-level semantic reconstruction, binding intra-class shared attributes via deconstruction. Mirroring human preference diversity in visual processing, where distinct individuals leverage dominant or contextual cues, we implement dominant and contextual consensus units to capture class-discriminative patterns and inherent distributional invariants, respectively. A consensus scheduler dynamically optimizes activation pathways, with final predictions emerging through multiplex consensus integration. Extensive evaluations across coarse- and fine-grained benchmarks demonstrate ConGCD's effectiveness as a consensus-aware paradigm. \href{https://github.com/lytang63/ConGCD}{\textcolor{purple}{Code is here}.}

\end{abstract}
\footnote{$^{*}$ Equal Contribution; $^{\dag}$ Corresponding Author.}
\section{Introduction}
\label{sec:intro}

\begin{figure}[t]
  \centering
   \includegraphics[width=0.95\linewidth]{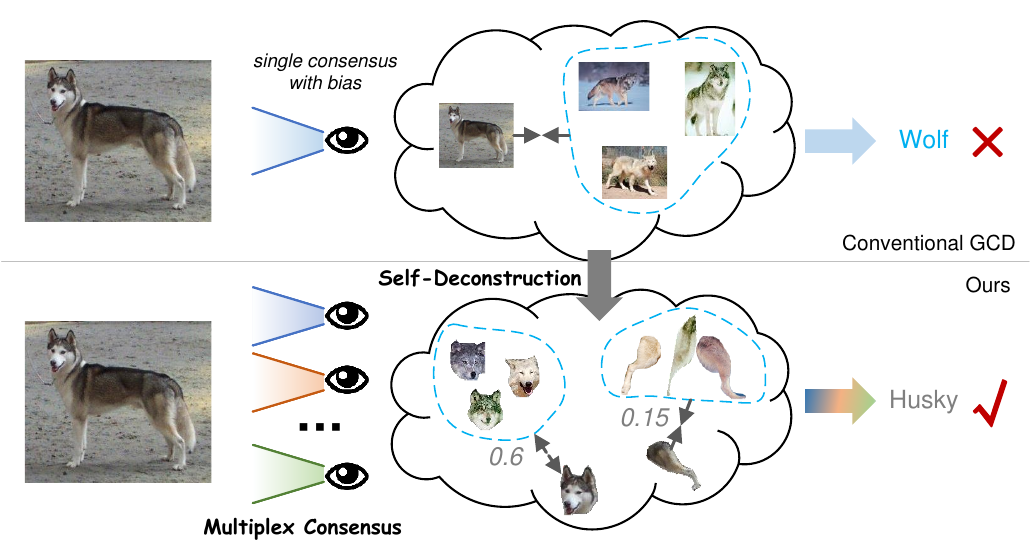}
   \caption{Generalized category discovery inherently demands multiplex consensus via self-deconstruction, rejecting image-level processing in favor of attribute-aware compositional reasoning.}
   \label{fig:clarify}
   \vspace{-4mm}
\end{figure}

Generalizing inductive and recognition capabilities to unknown categories of objects is an innate ability of human vision~\cite{chen1982topological}, but it is incredibly challenging for machine learning models. Generalized Category Discovery (GCD)~\cite{vaze2022generalized, wen2023parametric, zhao2023learning,rastegar2024selex,wang2024sptnet} is an advanced method to address this issue, which encourages models to cluster and recognize a subset of samples within the data that includes both unlabeled and unknown categories as well as known categories after training on partially labeled data.


From the perspective of representation learning~\cite{van2017neural}, current GCD paradigms including contrastive learning~\cite{vaze2022generalized,choi2024contrastive}, prototype classifiers~\cite{wen2023parametric}, and sample selection~\cite{ma2024active,rastegar2024selex} share a fundamental oversight: they process images as atomic entities, neglecting the compositional nature of visual recognition. For instance, CMS~\cite{choi2024contrastive} applies mean-shift to global embeddings, while SimGCD~\cite{wen2023parametric} distills category prototypes from holistic representations. Though effective for coarse discrimination, these methods fail to emulate the human capacity for attribute-level reasoning - the ability to decompose objects into semantic primitives and establish cross-category consensus through comparative analysis.

Humans can transfer previously acquired knowledge when learning new concepts~\cite{koh2020concept,zarlenga2022concept,tang2025ocrt}. If young children learn to distinguish between ``cows" and ``sheep" based on their shape, they can apply the same rule to differentiate between ``pigs" and ``horses". However, when it comes to telling ``pandas" and ``bears" apart, relying solely on shape can lead to confusion like ~\cref{fig:clarify}; attention to additional \textbf{visual primitives}~\cite{ferrari2007learning} such as stripes and ears is required. For instance, the Oxford Dictionary~\cite{dictionary1989oxford} defines a giraffe as ``\textit{a tall animal with a very long neck, long legs, and dark marks on its coat}". This definition involves two shape primitives (long neck, long legs) and one texture-color attribute (dark marks). Unfortunately, shortcut learning~\cite{geirhos2020shortcut} in deep networks typically activates a few primitives (such as the gray skin of an elephant~\cite{geirhos2018imagenettrained}), causing confusion between categories with similar primitives, like rhinoceroses.

\begin{figure*}[t]
  \centering
   \includegraphics[width=0.90\linewidth]{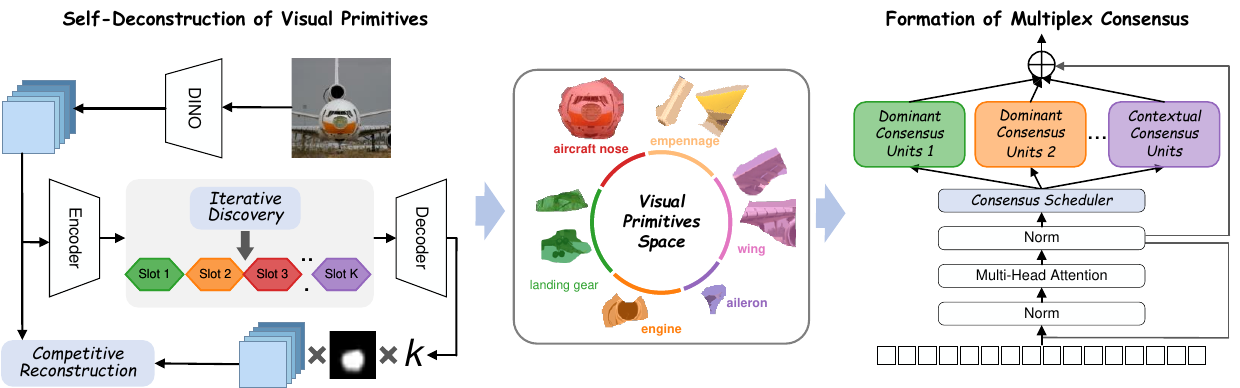}
   \caption{Overview of the proposed ConGCD: (1) Visual primitives are discovered iteratively via competitive reconstruction; (2) Each visual primitive is processed by dominant consensus units with primitive mask, while the inherent distribution is processed by contextual consensus units scheduler with holistic representation. The weight of each unit is determined by the consensus scheduler.}
   \label{fig:overview}
\end{figure*}

In this work, we propose cognitive science-inspired ConGCD, which attempts to address the above challenge by focusing on two aspects as shown in~\cref{fig:overview}: the \textbf{self-deconstruction of visual primitives} and the \textbf{formation of multiplex consensus}. (1) For the decoupling of visual primitives, we define a set of primitive-oriented representations to bind intra-class shared visual primitives to spatial regions~\cite{chen1982topological} in a construction manner. The competitive relationship among primitives-oriented representations ensures that each primitive mask focuses on a single visual attribute. Collectively, all primitive-oriented representations can be seen as a synthesis of the image's key visual primitives. (2) For the formation of multiplex consensus, unlike monolithic feature aggregation, we implement dominant and contextual consensus units: dominant units retain high-response neurons to capture class-discriminative patterns, while contextual units preserve weaker activations encoding distributional invariants. A dynamic consensus scheduler optimizes activation pathways between these units, with final predictions emerging through multiplex integration of their complementary perspectives. This process mimics how humans iteratively reconcile dominant cues (e.g., beak shape for birds) with contextual invariants (e.g., flight posture) when categorizing novel exemplars~\cite{krawczyk2012cognition, krawczyk2011hierarchy}.

Our main contributions can be summarized as follows:

\begin{itemize}
\item A mechanism inducing visual primitives via competitive binding, enabling unsupervised discovery of primitive-discriminative attributes through topological competition.
\item A multiplex consensus framework: dominant units retain high-response neurons for class-related variables, while contextual units preserve weak activations for inherent distributions, coordinated by dynamic scheduling.
\item Extensive validation showing our paradigm’s plug-and-play compatibility with GCD methods, achieving state-of-the-art performance across seven benchmarks.
\end{itemize}

\section{Related Works}
\label{sec:relat}

\subsection{Generalized Category Discovery}
\label{subsec:gcd}

In Generalized Category Discovery (GCD) \cite{vaze2022generalized,tang2025generalized}, the goal is to classify unlabeled samples into known and unknown classes using a limited labeled dataset from known classes, reflecting a realistic approach to real-world image recognition. Initially, GCD was formulated by \cite{vaze2022generalized}, employing contrastive and semi-supervised learning methods that relied heavily on clustering during inference. Consequently, SimGCD, a robust baseline approach that replaces clustering with a classifier, is trained using a pseudo-labeling strategy that has shown remarkable effectiveness in SSL \cite{wen2023parametric}. To address this limitation, subsequent works aim to improve feature representation by exploiting underlying cross-instance relationships \cite{pu2024federated}. An EM-like framework is proposed to alternate between contrastive learning and class number estimation \cite{zhao2023learning}. Recent works have proposed using parametric classifiers to avoid prediction biases \cite{elsayed2022savi++}. However, previous studies \cite{ma2024active, otholt2024guided} focused on class tokens for clustering, overlooking that similar visual attributes across samples can cause inter-class boundary confusion.


\subsection{Object-Centric Learning}
\label{subsec:ocl}


In the evolving landscape of Object-Centric Learning (OCL), Slot Attention established a significant milestone \cite{locatello2020object}. This mechanism enabled further research, like \cite{elsayed2022savi++}, which extended OCL to videos, using temporal consistency and dynamics within video sequences to better differentiate objects from backgrounds. Furthermore, to address the stability issue in single-view OCL, \cite{kim2023shepherding} introduces modules to guide slots away from background noise and towards consistent object representation. \cite{aydemir2023self} explores self-supervision, offering the first unsupervised multi-object segmentation in real-world videos. And the challenge of adapting OCL to complex real-world settings was tackled by \cite{seitzer2022bridging,tang2024bootstrap}, bridging the gap between lab and real-world scenarios. OCL currently concentrates on object-level separation \cite{li2024prompt}, and we extend it to a more fine-grained visual attribute discovery level, and let the model innately have the inductive ability for the semantics of visual attributes.

\section{Preliminaries}
\label{sec:pre}

\subsection{Visual Primitives}
\label{subsec:vs}

The generation process of an sample $\mathbf{x}$ is actually a combination process of a set of visual primitives $s^1, \cdots, s^K$ ($s^{1:K}$) with $s^i \in \mathcal{S}$, where $\mathcal{S}$ is a finite set~\cite{ferrari2007learning}. Considering the joint distribution of $\mathbf{x}$ and $s^{1:K}$, due to biases in observation and annotation and deviation from the real-world distribution, the label of a sample can depend on one or more primitives. The data generation process can be expressed as

\begin{equation}
\begin{aligned}
&\mathbf{z} \sim p(\mathbf{z}), \quad s^i \sim p(s^i|\mathbf{z}), \quad \mathbf{x} \sim p(\mathbf{x}|\mathbf{z}),\\
&p(s^{1:K},\mathbf{x}) = p(s^{1:K})\int p(\mathbf{x}|\mathbf{z})p(\mathbf{z}|s^{1:K})d\mathbf{z}, 
\end{aligned}
  \label{eq:vs}
\end{equation}
where $\mathbf{z}$ is a factor in the latent space. We believe that if different marginal distributions of given primitives exist but they share the same conditional generation process, there will be distribution shift between categories, that is $p_{\text{class-n}}(s^{1:K}) \neq p_{\text{class-n$^{_{'}}$}}(s^{1:K})$. However, due to the shared generative model between distributions, certain primitives subset $\tilde{\mathcal{A}}$ are highly similar across inter-class, we have $p_{\text{class-n}}(\tilde{s}^{1:\tilde{K}},\mathbf{x}) = p_{\text{class-n$^{_{'}}$}}(\tilde{s}^{1:\tilde{K}})\int p(\mathbf{x}|\mathbf{z}) 
 p(\mathbf{z}|\tilde{s}^{1:\tilde{K}})d\mathbf{z}$, where $\tilde{s}^i \in \tilde{\mathcal{S}}$ and $\tilde{\mathcal{S}} \subseteq \mathcal{S}$. The similarity of inter-class primitives is disastrous for distinguishing between categories. Therefore, we aim to find a complementary set $\mathcal{S} \setminus \tilde{\mathcal{S}}$, the discriminative visual primitives that maximize the distribution difference between categories, serving as the factors that contribute the most to inter-class separation.


\subsection{Problem Definition of GCD}
\label{subsec:pre_gcd}

we consider a labeled subset for each dataset, denoted as $\mathcal{D}_l = \{ (\mathbf{x}_i^l, y_i^l) \}\subset\mathcal{X}\times\mathcal{Y}_l$. Additionally, there is an unlabeled subset $\mathcal{D}_u = \{ (\mathbf{x}_i^u, y_i^u) \}\subset\mathcal{X}\times\mathcal{Y}_u$. In $\mathcal{D}_l$, only known classes are present, i.e., $\mathcal{Y}_l = \mathcal{C}_{known}$. On the other hand, $\mathcal{D}_u$ contains both known and novel classes, $\mathcal{Y}_u = \mathcal{C}_{known} \cup \mathcal{C}_{novel}$. The task for models is to perform clustering on both the known and novel classes in $\mathcal{D}_u$. The number of novel classes, represented as $N_{novel}$, can be determined in advance~\cite{vaze2022generalized, pu2023dynamic, zhao2023learning}. The functions $f(\cdot)$ and $g(\cdot)$ act as the feature extractor and projection head respectively. Both the feature $\mathbf{h}_i = f(\mathbf{x}_i)$ and the projected embedding $\mathbf{z}_i = g(\mathbf{h}_i)$ are subjected to L-2 normalization.

\section{Methodology}
\label{sec:method}

\subsection{Self-Deconstruction of Visual Primitives}
\label{subsec:decoup_vs}

\textbf{Visual primitives Iterative Discovery.} 
Referring to the process in which humans compare the visual features of an object with the existing knowledge in the brain and conduct cyclic searches after receiving visual signals~\cite{chen1982topological}, we define visual primitive discovery as an iterative process. It originates from slot-attention~\cite{locatello2020object}. We predefine primitive-centric representations (learnable embeddings), $\mathbf{s} \sim \mathcal{N}(\mathbf{s} ; \boldsymbol{\mu}, \boldsymbol{\sigma}) \in \mathbb{R}^{K \times D_s}$. $K$ is the number of visual primitives, and $D_s$ is the number of dimensions.

For the pre-trained encoder's output $\mathbf{z} = g(f(\mathbf{x})) \in \mathbb{R}^{N \times D_z}$, the attention mechanism of the human-like perception process~\cite{vaswani2017attention} is implemented. Representing $\mathcal{K}_\beta$, $\mathcal{Q}_\gamma$ and $\mathcal{V}_\phi$ for $\mathbf{z}$ as the projection networks of key, query, and value with parameters $\beta$, $\gamma$, and $\phi$, respectively. The \underline{dis}covery process of \underline{v}isual \underline{p}rimitives $\operatorname{vpdis(\cdot)}$ and the attention function $\operatorname{attn(\cdot)}$ can be defined as

\begin{equation}
\begin{aligned}
\operatorname{vpdis}(\boldsymbol{A}, \mathbf{v})&=\boldsymbol{A}^T \mathbf{v}, \quad A_{i j}=\frac{\operatorname{attn}(\mathbf{q}, \mathbf{k})_{i j}}{\sum_{l=1}^K \operatorname{attn}(\mathbf{q}, \mathbf{k})_{l j}}, \\
\operatorname{attn}(\mathbf{q}, \mathbf{k})&=\frac{e^{M_{i j}}}{\sum_{l=1}^N e^{M_{i l}}}, \quad \boldsymbol{M}=\frac{\mathbf{k q}^T}{\sqrt{D_s}},
\end{aligned}
  \label{eq:attn}
\end{equation}

Where $\boldsymbol{A}$ is the attention matrix. The $\mathbf{k}=\mathcal{K}_\beta(\mathbf{z}) \in \mathbb{R}^{N \times D_s}$ and $\mathbf{v}=\mathcal{V}_\phi(\mathbf{z}) \in \mathbb{R}^{N \times D_s}$ are the key and value vectors, respectively. To ensure the binding of the attention relationship between visual primitives and feature embeddings, queries are a function of visual primitives $\mathbf{s}$, and the formal expression is $\mathbf{q}=\mathcal{Q}_\gamma(\mathbf{s}) \in \mathbb{R}^{K \times D_s}$.

Similar to the process of thinking and reasoning~\cite{alexander2016relational}, visual primitives are iteratively refined over $T$ attention iterations. Specifically, in iteration $t$, the queries are given by $\hat{\mathbf{q}}^t=\mathcal{Q}_\gamma\left(\mathbf{s}^t\right)$. The Gated Recurrent Unit (GRU)~\cite{chung2014empirical,dey2017gate} is applied, denoted as $\mathcal{G}_\theta$. The visual primitives update process can be summarized as $\mathbf{s}^{t+1}=\operatorname{vpdis}\left(\operatorname{attn}\left(\hat{\mathbf{q}}^t, \mathbf{k}\right), \mathbf{v}\right)$.


Given $\mathbf{x}$ at $t = T$ iteration, the posterior distribution of the visual primitives is

\begin{equation}
\begin{aligned}
&p\left(\mathbf{s}^T \mid \mathbf{x}\right) = \\
&\delta\left(\mathbf{s}^T-\prod_{t=1}^T \mathcal{G}_\theta\left(\mathbf{s}^{t-1}, \operatorname{vpdis}\left(\operatorname{attn}\left(\hat{\mathbf{q}}^{t-1}, \mathbf{k}\right), \mathbf{v}\right)\right)\right) ,
\end{aligned}
\end{equation}
where $\delta(\cdot)$ represents the Dirac delta distribution~\cite{belloni2014infinite}, we sample s from a Gaussian distribution~\cite{goodman1963statistical}, at $t = 0$ iteration for initialization so that the distribution of $\mathbf{s}^{0}$ represents relatively uniform visual primitives, preventing biases in the visual primitive discovery process.

\noindent

\textbf{Deconstruction via Competitive Reconstruction.}
$\mathbf{\hat{z}}=\Phi_d\left(\mathbf{s}_k\right)$
The discovery process of visual primitives is unsupervised, and we complete it through the surrogate task of reconstruction. After $\mathbf{s}^{T}$ goes through the above iterative discovery, it is sent in as a general lightweight multi-layer perceptron (MLP) as the decoder~\cite{seitzer2023bridging}, denoted as $\Phi_d(\cdot)$. $\Phi_d(\cdot)$ spatially broadcasts $K$ visual primitives to the same spatial dimension as $\mathbf{\hat{z}}_k=\Phi_d\left(\mathbf{s}_k\right) \in \mathbb{R}^{N \times (D_z+1)}$. The reconstructed $\mathbf{\hat{z}}_k$ is composed of $D_z+1$ channels. After channel division, we obtain the reconstructed feature embeddings $\mathbf{\hat{z}}_k \in \mathbb{R}^{N \times D_s}$ and the activation region ${\alpha}_k \in \mathbb{R}^{N \times 1}$. In order to bind the visual primitive embedding to the region with the same semantics in the image, we use $\operatorname{softmax}$ on ${\alpha}_k$ for normalization over $K$ visual primitives to introduce the competitive relationship between visual primitives, so that a certain primitive is associated with certain pixels

\begin{equation}
\mathbf{\hat{z}}=\sum_{k=1}^K \mathbf{\hat{z}}_k \odot \boldsymbol{m}_k, \quad \boldsymbol{m}_k=\underset{k}{\operatorname{softmax}}({\alpha}_k) .
\end{equation}

The optimization objective of the iterative discovery and decoupling process of visual primitives is simple

\begin{equation}
\mathcal{L}_{\text {rec}}=\|\mathbf{\hat{z}}-\mathbf{z}\|^2, \quad \mathbf{\hat{z}}=\Phi_d(\mathbf{s}) .
\end{equation}

This self-supervised deconstruction establishes basic visual primitives through competitive binding, enabling machines to mirror human attribute-level perception. By disentangling objects into composable factors, we equip models with intrinsic interpretability for fine-grained comparisons while maintaining adaptability to novel category structures.

\begin{table*}[!t]
  \centering
  \scalebox{0.85}{
    \begin{tabular}{c|l|ccc|ccc|ccc|ccc}
      \toprule
      & \multirow{2}{*}{\textbf{Method}}
      & \multicolumn{3}{c}{\textbf{CUB-200}}
      & \multicolumn{3}{c}{\textbf{FGVC-Aircraft}}
      & \multicolumn{3}{c}{\textbf{Stanford-Cars}}
      & \multicolumn{3}{c}{\textbf{Average}} \\ 
      \cmidrule(lr){3-5} \cmidrule(lr){6-8} \cmidrule(lr){9-11} \cmidrule(lr){12-14}
      & & All & Known&Novel&All&Known&Novel&All&Known&Novel&All&Known&Novel \\
      \midrule
      \multirow{16}{*}{\rotatebox{90}{DINOv1}} & GCD \cite{vaze2022generalized} &51.3&56.6&48.7&45.0&41.1&46.9&39.0&57.6&29.9&45.1&51.8&41.8 \\
      & GPC \cite{zhao2023learning}&52.0&55.5&47.5&43.3&40.7&44.8&38.2&58.9&27.4&44.5&51.7&39.9 \\
      & XCon \cite{fei2022xcon}&52.1&54.3&51.0&47.7&44.4&49.4&40.5&58.8&31.7&46.8&52.5&44.0 \\
      & PromptCAL \cite{zhang2023promptcal} &62.9&64.4&62.1&52.2&52.2&52.3&50.2&70.1&40.6&55.1&62.2&51.7 \\
      & AMEND \cite{banerjee2024amend}&64.9&75.6&59.6&52.8&61.8&48.3&56.4&73.3&48.2&58.0&70.2&52.0 \\
      & $\mu$GCD \cite{vaze2024no}&65.7&68.0&64.6&53.8&55.4&53.0&56.5&68.1&\textbf{50.9}&58.7&63.8&56.2 \\
      & CMS \cite{choi2024contrastive}&68.2&76.5&64.0&56.0&63.4&52.3&56.9&76.1&47.6&60.4&72.0&54.6 \\
      & InfoSieve \cite{rastegar2024learn}&69.4&77.9&65.2&56.3&63.7&52.5&55.7&74.8&46.4&60.5&72.1&54.7 \\
      \cmidrule{2-14}
      & SimGCD \cite{wen2023parametric}&60.3&65.6&57.7&54.2&59.1&51.8&53.8&71.9&45.0&56.1&65.5&51.5 \\
      & \cellcolor{gray!15} \quad + ConGCD&\cellcolor{gray!15}61.6&\cellcolor{gray!15}65.1&\cellcolor{gray!15}59.5&\cellcolor{gray!15}55.0&\cellcolor{gray!15}58.4&\cellcolor{gray!15}53.1&\cellcolor{gray!15}54.5&\cellcolor{gray!15}72.2&\cellcolor{gray!15}47.8&\cellcolor{gray!15}57.1&\cellcolor{gray!15}65.2&\cellcolor{gray!15}53.5 \\
      & LegoGCD \cite{cao2024solving}&63.8&71.9&59.8&55.0&61.5&51.7&57.3&75.7&48.4&58.7&69.7&53.3 \\
      & \cellcolor{gray!15} \quad + ConGCD&\cellcolor{gray!15}65.3&\cellcolor{gray!15}72.6&\cellcolor{gray!15}61.7&\cellcolor{gray!15}55.6&\cellcolor{gray!15}61.2&\cellcolor{gray!15}54.0&\cellcolor{gray!15}58.1&\cellcolor{gray!15}75.9&\cellcolor{gray!15}49.5&\cellcolor{gray!15}59.7&\cellcolor{gray!15}69.9&\cellcolor{gray!15}55.1 \\
      & SPTNet \cite{wang2024sptnet}&65.8&68.8&65.1&59.3&61.8&58.1&\underline{59.0}&\textbf{79.2}&49.3&61.4&69.9&57.5 \\
      & \cellcolor{gray!15} \quad + ConGCD&\cellcolor{gray!15}68.1&\cellcolor{gray!15}68.5&\cellcolor{gray!15}67.8&\cellcolor{gray!15}59.7&\cellcolor{gray!15}61.3&\cellcolor{gray!15}\textbf{59.2}&\cellcolor{gray!15}\textbf{59.1}&\cellcolor{gray!15}\underline{79.0}&\cellcolor{gray!15}\underline{49.8}&\cellcolor{gray!15}62.3&\cellcolor{gray!15}69.6&\cellcolor{gray!15}58.9 \\
      & SelEx \cite{rastegar2024selex}&\underline{78.7}&\textbf{81.3}&\underline{77.5}&\underline{60.9}&\textbf{70.3}&56.2&57.0&77.4&47.2&\underline{65.5}&\textbf{76.3}&\underline{60.3} \\
      & \cellcolor{gray!15} \quad + ConGCD&\cellcolor{gray!15}\textbf{81.7}&\cellcolor{gray!15}\underline{80.4}&\cellcolor{gray!15}\textbf{82.4}&\cellcolor{gray!15}\textbf{62.5}&\cellcolor{gray!15}\underline{70.2}&\cellcolor{gray!15}\underline{58.7}&\cellcolor{gray!15}57.5&\cellcolor{gray!15}77.5&\cellcolor{gray!15}47.9&\cellcolor{gray!15}\textbf{67.3}&\cellcolor{gray!15}\underline{76.0}&\cellcolor{gray!15}\textbf{63.0} \\ 
      \cmidrule{2-14}
      & Avg. $\triangle$ &\textcolor{ForestGreen}{\textbf{+2.03}}&\textcolor{red!50!white}{-0.25}&\textcolor{ForestGreen}{\textbf{+2.83}}&\textcolor{ForestGreen}{\textbf{+0.85}}&\textcolor{red!50!white}{-0.40}&\textcolor{ForestGreen}{\textbf{+1.80}}&\textcolor{ForestGreen}{\textbf{+0.53}}&\textcolor{ForestGreen}{\textbf{+0.10}}&\textcolor{ForestGreen}{\textbf{+1.28}}&\textcolor{ForestGreen}{\textbf{+1.18}}&\textcolor{red!50!white}{-0.18}&\textcolor{ForestGreen}{\textbf{+1.98}} \\
      \midrule
      \multirow{5}{*}{\rotatebox{90}{DINOv2}} & GCD \cite{vaze2022generalized}&71.9&71.2&72.3&55.4&47.9&59.2&65.7&67.8&64.7&64.3&62.3&65.4 \\
      & SimGCD\cite{wen2023parametric}&71.5&78.1&68.3&63.9&69.9&60.9&71.5&81.9&66.6&69.0&76.6&65.3 \\
      & $\mu$GCD \cite{vaze2024no}&74.0&75.9&73.1&66.3&68.7&65.1&76.1&91.0&68.9&72.1&78.5&69.0 \\
      & SelEx\cite{rastegar2024selex}&\underline{86.0}&\underline{86.5}&\underline{85.7}&\textbf{82.1}&\textbf{84.8}&\underline{80.7}&\textbf{80.5}&\underline{91.8}&\textbf{75.0}&\textbf{82.9}&\underline{87.7}&\textbf{80.5} \\
      & \cellcolor{gray!15} \quad + ConGCD&\cellcolor{gray!15}\textbf{86.3}&\cellcolor{gray!15}\textbf{87.4}&\cellcolor{gray!15}\textbf{85.8}&\cellcolor{gray!15}\underline{81.7}&\cellcolor{gray!15}\underline{83.3}&\cellcolor{gray!15}\textbf{81.0}&\cellcolor{gray!15}\underline{79.8}&\cellcolor{gray!15}\textbf{93.1}&\cellcolor{gray!15}\underline{73.3}&\cellcolor{gray!15}\underline{82.6}&\cellcolor{gray!15}\textbf{87.9}&\cellcolor{gray!15}\underline{80.1} \\
      \bottomrule
    \end{tabular}
  }
  \caption{Results on the \textbf{fine-grained} semantic shift benchmark. ConGCD offers excellent cross-scheme and cross-model compatibility.}
  \label{tab:fine}
\end{table*}

\subsection{Formation of Multiplex Consensus}
\label{subsec:moae}


Models need to understand the visual primitives. However, exclusive reliance on dominant feature activations~\cite{geirhos2018imagenettrained} creates self-reinforcing biases that degrade unknown-class generalization, as shown in~\cref{tab:ablation}, where strongly activated neurons encode class-agnostic structural primitives, while their weakly activated counterparts preserve cross-category contextual patterns~\cite{li2024emergence}.

\begin{figure}[ht]
  \centering
   \includegraphics[width=0.95\linewidth]{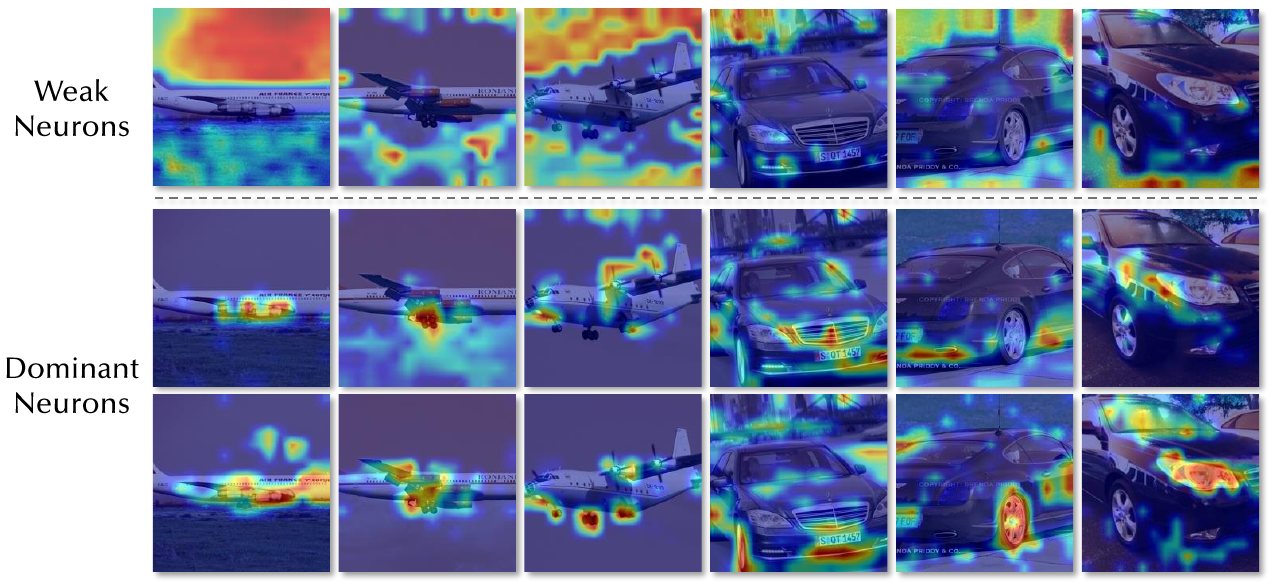}
   \caption{Dominant and weak neurons capture the class-related variables and inherent distributions, respectively.}
   \label{fig:gradcam}
\end{figure}

This duality (as visualized in~\cref{fig:gradcam}) drives our consensus formation through two complementary pathways: (1) Dominant Consensus Units distill class-discriminative knowledge by routing high-activation visual primitives through neural competition layers, preventing primitive overcommitment to known categories. (2) Contextual Consensus Units continuously assimilate weakly activated embeddings to model class-transcendent relationships. The resultant multiplex architecture achieves self-stabilization through consensus scheduler, where primitive specificity and contextual generality co-evolve.

\noindent

\textbf{Dominant Consensus Units.} 
For the output $\mathbf{z}^i \in \mathbb{R}^{N \times D^i_z}$ of the $i$-th block in the network forward process, in order to be consistent with the pixel-level relationship of visual primitives, we selectively retain the top $r$ (\%) of neuron activation responses at the token-level from the spatial dimension. Specifically, we customize a sorting function $\operatorname{sort}(\cdot)$ from large to small for neurons to generate $\mathbf{z}_{attr}^i$ after filtering neuron activation values. Furthermore, for each consensus units, their attention should be focused on a single visual primitive to ensure the expertise of the units (Analogous to human visual preferences, some people rely on the head to distinguish animals, and some people rely on the tail). We directly use $\boldsymbol{m}_k$ to represent the activation of the visual primitive region to limit the input of $k$-th unit. The forward process of each dominant consensus unit is

\begin{equation}
\mathbf{z}_{Dom}^{ik}:=\left\{\begin{aligned}
&\mathbf{z}^i \odot \boldsymbol{m}_k,  \text { if } \mathbf{z}^i[:, :] \geq \operatorname{Top-R}_{k}(\operatorname{sort}(\mathbf{z}^i[:,:])) \\
&0,  \text { otherwise },
\end{aligned}\right.
\label{eq:attr}
\end{equation}
where $R_{k}=N \times r_{k}$ in $\operatorname{Top-R_{k}}(\cdot)$, for visual primitive schedulers used to capture class-related variables, they need to perceive highly activated neurons and provide features conducive to separation between classes.

\noindent

\textbf{Contextual Consensus Units.} 
Due to the inherent distribution of the dataset, only maximizing the extraction of class-related visual primitives will cause difficulties for the model in recognizing novel classes. Therefore, we propose contextual consensus units based on weak neurons. It accepts the complete $\mathbf{z}^i$ as input and perceives all visual primitives of the object. We decompose the low-activation region in $\mathbf{z}^i$

\begin{equation}
\mathbf{z}_{Con}^{i}:=\left\{\begin{aligned}
&\mathbf{z}^i,  \text { if } \mathbf{z}^i[:, :] \leq \operatorname{Top-R}(\operatorname{sort}(\mathbf{z}^i[:,:])) \\
&0,  \text { otherwise } ,
\end{aligned}\right.
\label{eq:holi}
\end{equation}
which is similarly to~\cref{eq:attr}, but dominant consensus units on class-independent features activated in weak neurons, aiming to extract the information common to each sample from the feature embedding itself. This prevents the model from overly focusing on the visual primitives of known classes and from effectively perceiving novel classes.

\noindent

\textbf{Consensus Scheduler.} 
The self-deconstruction process manifests in ViT blocks~\cite{caron2021emerging} through dual activation polarization: while multi-head self-attention (MHA)~\cite{vaswani2017attention} captures cross-patch dependencies, the decomposed dominant/contextual consensus pathways exhibit complementary neuron activation patterns. We reformulate the FFN as parallel consensus units governed by activation separation:

\begin{equation}
\begin{aligned}
    &f_{\text{ConGCD}}(\mathbf{z})=\sum_{k=1}^{K+M} G(\mathbf{z})_k \cdot E_k(\mathbf{z}_{Dom}^{k} / \mathbf{z}_{Con}) \\
    &=\sum_{k=1}^{K+M} \operatorname{softmax}(\boldsymbol{W} \mathbf{z}) \cdot \boldsymbol{W}_{\mathrm{FFN}_k}^2 \phi(\boldsymbol{W}_{\mathrm{FFN}_k}^1 \mathbf{z}_{Dom}^{k} / \mathbf{z}_{Con}) , 
\end{aligned}
\end{equation}
where $\boldsymbol{W}$ and $\boldsymbol{W}_{\mathrm{FFN}_k}^{1/2}$ are the learnable parameters. $\phi(\cdot)$ is a nonlinear activation function. The primitive scheduler operates densely for $K$ dominant units and $M$ contextual units and assigns a non-zero probability to each unit.



Due to inconsistent preferences across units, determining the ratio $r_{k}$ (\%) for detecting dominant/contextual activation neurons becomes challenging. We address this via an Activation Separation Distributor (ASR) that dynamically generates unit-specific separation ratios from $\mathbf{z}$, enabling adaptive boundary determination for activation decomposition per sample. Scheduler adopt a new $R^{*}$ to screen neurons using $\operatorname{Top-R^{*}_{k}}(\cdot)$:

\begin{equation}
R^{*}_{k}=N  \times\left(r_{k} + \eta \cdot  \operatorname{softmax}(\boldsymbol{W}_{\mathrm{ASR}} \mathbf{z})[k] \right),
\end{equation}
where $W_{\text{ASR}}$ is learnable, $\eta$ controls scaling, and $N$ denotes patch count. This allows sample-aware adaptation to diverse visual primitives while maintaining activation balance between classes.In general, given $\mathbf{z}$ as the input, the update process is

\begin{equation}
\mathbf{z}_{\text {out}}=f_{\text{ConGCD}}(\operatorname{LN}(\mathbf{z}))+\mathbf{z},
\end{equation}
where $\operatorname{LN}(\cdot)$ indicates layer normalization, and $\mathbf{z}_{\text{out}}$ is the output and is sent to the next block. The multiplex consensus framework resolves fundamental tension in GCD through symbiotic coordination, dominant units sharpen categorical boundaries while contextual units maintain open-world continuity, mirroring human vision's dual processing of focal attributes and environmental context.

\begin{table*}[t]
  \centering
  \scalebox{0.90}{
\setlength{\tabcolsep}{3pt}
\renewcommand{\arraystretch}{1.1}
\begin{tabular}{c|l|ccc|ccc|ccc|ccc}
\toprule
&\multirow{2}{*}{\textbf{Method}}
&\multicolumn{3}{c}{\textbf{CIFAR-10}}
&\multicolumn{3}{c}{\textbf{CIFAR-100}}
&\multicolumn{3}{c}{\textbf{ImageNet-100}}
&\multicolumn{3}{c}{\textbf{Average}}\\
\cmidrule(lr){3-5} \cmidrule(lr){6-8} \cmidrule(lr){9-11} \cmidrule(lr){12-14}
& &All & Known & Novel & All & Known & Novel & All & Known & Novel & All & Known & Novel\\
\midrule
\multirow{17}{*}{\rotatebox{90}{DINOv1}}
&ORCA \cite{2021Open}         & 96.9 & 95.1 & 97.8 & 74.2 & 82.1  & 67.2&79.2	&93.2	&72.1&83.4&90.1&79.0\\
&GCD \cite{vaze2022generalized}    & 91.5 & \textbf{97.9} & 88.2 & 73.0 & 76.2  & 66.5&74.1 & 89.8&	66.3&79.5 & 88.0  &73.7 \\
&GPC \cite{zhao2023learning} &90.6& 97.6& 87.0& 75.4 &84.6& 60.1&75.3& 93.4 & 66.7&80.4 &91.9& 71.3\\
&XCon \cite{fei2022xcon} & 96.0&97.3&95.4&74.2&81.2&60.3&77.6& 93.5 &69.7&82.6&90.7&75.1\\
&PIM \cite{chiaroni2023parametric}       & 94.7 & 97.4 & 93.3 & 78.3 & 84.2 & 66.5&83.1&\textbf{95.3}&	77.0&85.4 & \textbf{92.3} & 78.9\\
&PromptCAL \cite{zhang2023promptcal} &\textbf{97.9} & 96.6 & \underline{98.5} & 81.2 & 84.2 & 75.3&83.1	&92.7	&78.3 &87.4 & 91.2 & 84.0\\
&DCCL \cite{pu2023dynamic}        & 96.3 & 96.5 & 96.9 & 75.3 & 76.8  & 70.2&80.5&90.5	&76.2&84.0 & 87.9  & 81.1\\
&InfoSieve \cite{rastegar2024learn}                  &94.8&\underline{97.7} &93.4
& 78.3& 82.2 &70.5& 80.5 &93.8 &73.8&84.5& 91.2 &79.2\\
\cmidrule(lr){2-14}
&SimGCD \cite{wen2023parametric}       & 97.1 & 95.1 &98.1 &80.1 & 81.2 & 77.8& 83.0 &93.1&77.9&86.7 & 89.8& 84.6\\
& \cellcolor{gray!15}\quad + ConGCD
& \cellcolor{gray!15}97.3&\cellcolor{gray!15}95.1&\cellcolor{gray!15}98.4
& \cellcolor{gray!15}81.3&\cellcolor{gray!15}82.5&\cellcolor{gray!15}78.9
& \cellcolor{gray!15}83.5&\cellcolor{gray!15}92.8&\cellcolor{gray!15}78.6
& \cellcolor{gray!15}87.4&\cellcolor{gray!15}90.1&\cellcolor{gray!15}85.3\\

&LegoGCD \cite{cao2024solving}&97.1&94.3&\underline{98.5}&81.8&81.4&\textbf{82.5}&\textbf{86.3}&\underline{94.5}&\underline{82.1}&\underline{88.4}&90.1&\textbf{87.7}\\
& \cellcolor{gray!15}\quad + ConGCD
& \cellcolor{gray!15}97.2&\cellcolor{gray!15}94.8&\cellcolor{gray!15}97.9
& \cellcolor{gray!15}\underline{82.2}&\cellcolor{gray!15}82.1&\cellcolor{gray!15}\underline{82.3}
& \cellcolor{gray!15}85.8&\cellcolor{gray!15}94.4&\cellcolor{gray!15}81.7
& \cellcolor{gray!15}\underline{88.4}&\cellcolor{gray!15}90.4&\cellcolor{gray!15}\underline{87.3}\\

&SPTNet \cite{wang2024sptnet} &97.3&95.0&\textbf{98.6}& 81.3&84.3&75.6&85.4&93.2&81.4&88.0&90.8&85.2\\
& \cellcolor{gray!15}\quad + ConGCD
& \cellcolor{gray!15}\underline{97.4}&\cellcolor{gray!15}95.2&\cellcolor{gray!15}\underline{98.5}
& \cellcolor{gray!15}\textbf{82.5}&\cellcolor{gray!15}\textbf{85.9}&\cellcolor{gray!15}77.3
& \cellcolor{gray!15}\underline{85.9}&\cellcolor{gray!15}93.4&\cellcolor{gray!15}\textbf{82.5}
& \cellcolor{gray!15}\textbf{88.6}& \cellcolor{gray!15}91.5&\cellcolor{gray!15}86.1\\

&SelEx \cite{rastegar2024selex}& 94.1&\underline{97.7}&92.2&80.0&84.8&70.4&82.3& 93.9 &76.5&85.4& 92.1&79.7\\
& \cellcolor{gray!15}\quad + ConGCD
& \cellcolor{gray!15}95.7&\cellcolor{gray!15}97.6&\cellcolor{gray!15}94.8
& \cellcolor{gray!15}\underline{80.5}&\cellcolor{gray!15}\underline{84.9}&\cellcolor{gray!15}71.7
& \cellcolor{gray!15}83.5& \cellcolor{gray!15}94.2&\cellcolor{gray!15}78.0
& \cellcolor{gray!15}86.6& \cellcolor{gray!15}\underline{92.2}&\cellcolor{gray!15}81.5\\
\cmidrule(lr){2-14}
& Avg. $\triangle$ & 
\textbf{\textcolor{ForestGreen}{+0.50}} & 
\textbf{\textcolor{ForestGreen}{+0.15}} & 
\textbf{\textcolor{ForestGreen}{+0.55}} & 
\textbf{\textcolor{ForestGreen}{+0.83}} & 
\textbf{\textcolor{ForestGreen}{+0.93}} & 
\textbf{\textcolor{ForestGreen}{+0.98}} & 
\textbf{\textcolor{ForestGreen}{+0.43}} & 
\textcolor{red!50!white}{-0.03} & 
\textbf{\textcolor{ForestGreen}{+0.73}} & 
\textbf{\textcolor{ForestGreen}{+0.63}} & 
\textbf{\textcolor{ForestGreen}{+0.35}} & 
\textbf{\textcolor{ForestGreen}{+0.75}} \\
\bottomrule
\end{tabular}
}
\caption{Results on the \textbf{coarse-grained} classification benchmark. ConGCD offers excellent cross-scheme and cross-model compatibility.}
\label{tab:coarse}
\end{table*}

\section{Experiments}
\label{sec:exp}

Through comprehensive experimental verification and analysis, we aim to answer the following questions. (1) Can ConGCD improve the accuracy of GCD? (2) Can ConGCD be plugged and played in existing GCD schemes? (3) What is the source of the performance improvement of ConGCD? (4) Is ConGCD robust to changes in hyperparameters?

\subsection{Setup}
\label{subsec:exp_setup}

\textbf{Benchmarks.} ConGCD is evaluated on a total of seven image recognition benchmarks, namely four fine-grained datasets, CUB-200-2011~\cite{wah2011caltech}, Stanford Cars~\cite{krause20133d}, FGVC Aircraft~\cite{maji2013fine}, Herbarium19~\cite{tan2019herbarium}, and three coarse-grained datasets, CIFAR10, CIFAR100~\citep{krizhevsky2009learning} and ImageNet100~\citep{geirhos2018imagenet}. When dealing with CUB, Stanford Cars, and FGVC Aircraft, to segregate target classes into sets of known and unknown, we follow the splits defined by the Semantic Shift Benchmark (SSB)~\cite{vaze2021open}. For the remaining datasets, the splits from the previous study~\cite{vaze2022generalized} are employed. Under the CIFAR100 benchmark, we designate 80\% of the classes as known. For the other benchmarks, the proportion of known classes is 50\%. Our labeled set, known as $\mathcal{D}_{l}$, consists of 50\% images from the known classes for all benchmarks.

\noindent
\textbf{Evaluation Protocols.} Following existing research \cite{rastegar2024selex}, we use Balanced Semi-Supervised K-means \cite{vu2010active} to cluster the entire image collection $\mathcal{D}$ and calculate the model's recognition accuracy on the dataset $\mathcal{D}_{u}$, which does not have access to ground truth labels. The Hungarian algorithm \cite{wright1990speeding} is used to obtain the optimal mapping relationship between the unlabeled cluster set and the ground truth labels. Instances in the unlabeled set $\mathcal{D}_{u}$ are divided into \textbf{\textit{Known}} and \textbf{\textit{Novel}} categories based on their category primitives for clustering accuracy evaluation. In addition, the accuracy of \textbf{\textit{All}} classes is reported, as it represents the comprehensive performance of the model.

\noindent
\textbf{Implementation Details.} We employ the pre-trained DINOv1 \cite{caron2021emerging} on ImageNet-1K \cite{krizhevsky2012imagenet} as image encoder, and keep it frozen. Features from the $10$-th block are utilized for the decoupling of visual primitives. We set number of visual primitives to 8. The Slot-Attention is trained with the Adam optimizer for 400 epochs with learning rate of 0.0005, batch size of 64. We construct the ConGCD for the last two blocks, following with the previos research \cite{rastegar2024selex,vaze2022generalized}. The \(\lambda\) = 0.35, label smoothing coefficient \(\alpha\) set to 0.5 in the fine-grained dataset and 0.1 in the coarse-grained dataset. Additionally, we conduct the above training on DINOv2 \cite{oord2018representation} pre-trained on ImageNet-22K \cite{ridnik2021imagenet}. All experiments are conducted on RTX-4090, and random number seeds are fixed at each stage to avoid data label leaks.

\subsection{Baselines}
\label{subsec:baselines}


In addition to the contrast-based method CMS \cite{choi2024contrastive} and the use of prototype classifiers in SimGCD \cite{wen2023parametric}, we also compare ConGCD with several other state-of-the-art existing methods. LegoGCD \cite{cao2024solving} \cite{vaze2022generalized} leverages a modular approach to discover novel categories in unlabelled datasets by combining different components. SPTNet \cite{wang2024sptnet} iteratively optimizes both model parameters and data parameters through spatial prompt tuning. SelEx \cite{rastegar2024selex} addresses GCD through self-expertise strategies for hierarchical pseudo-label.

\begin{figure*}[!th]
  \centering
  \subfloat[$log(\operatorname{rank}(\mathcal{A}))$ of fine-grained datasets]
  {\includegraphics[width=0.225\textwidth]{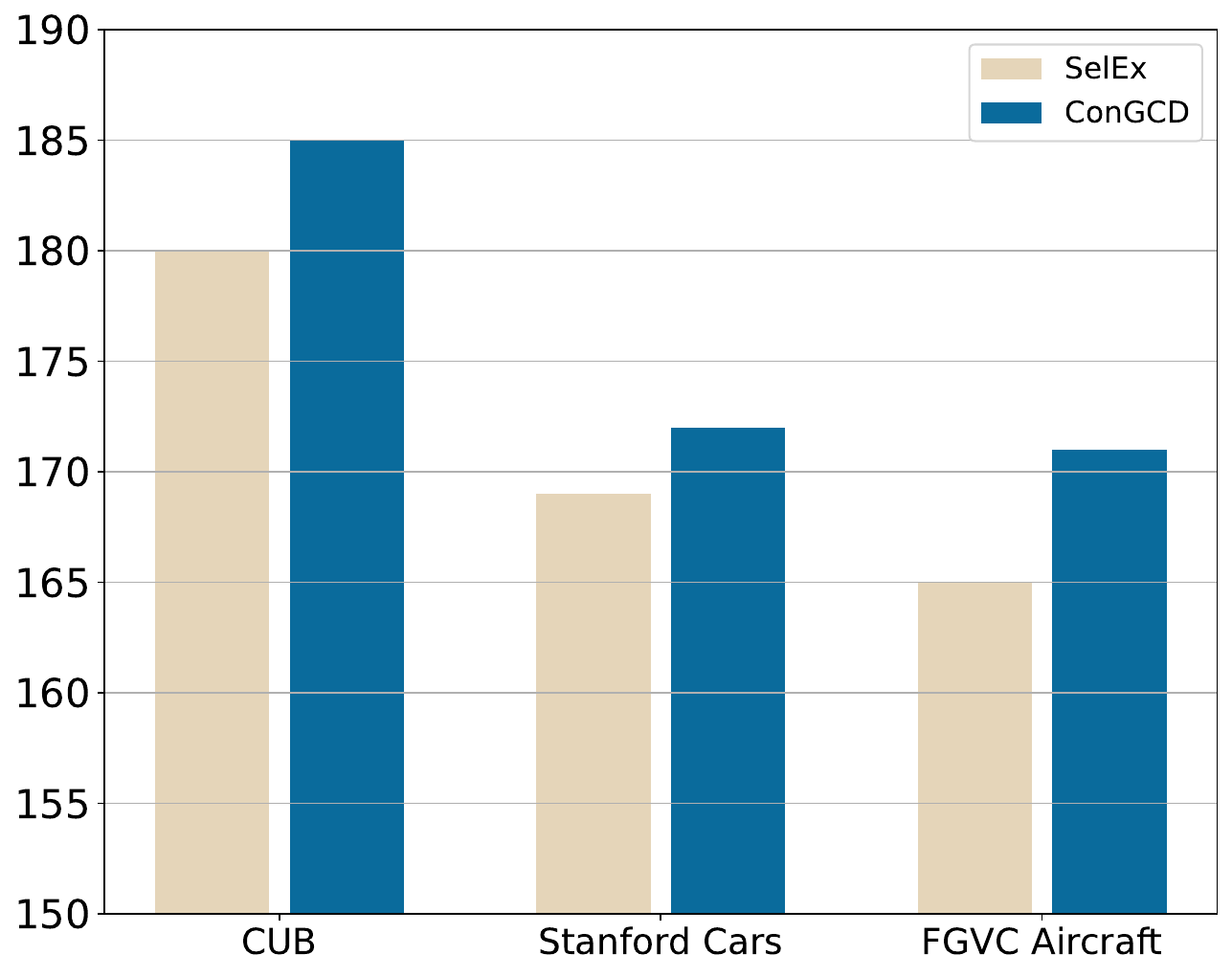}\label{fig:figure_von_log_fine}}
  \quad     
  \subfloat[$\hat{H}(\mathcal{A})$ of fine-grained datasets]
  {\includegraphics[width=0.22\textwidth]{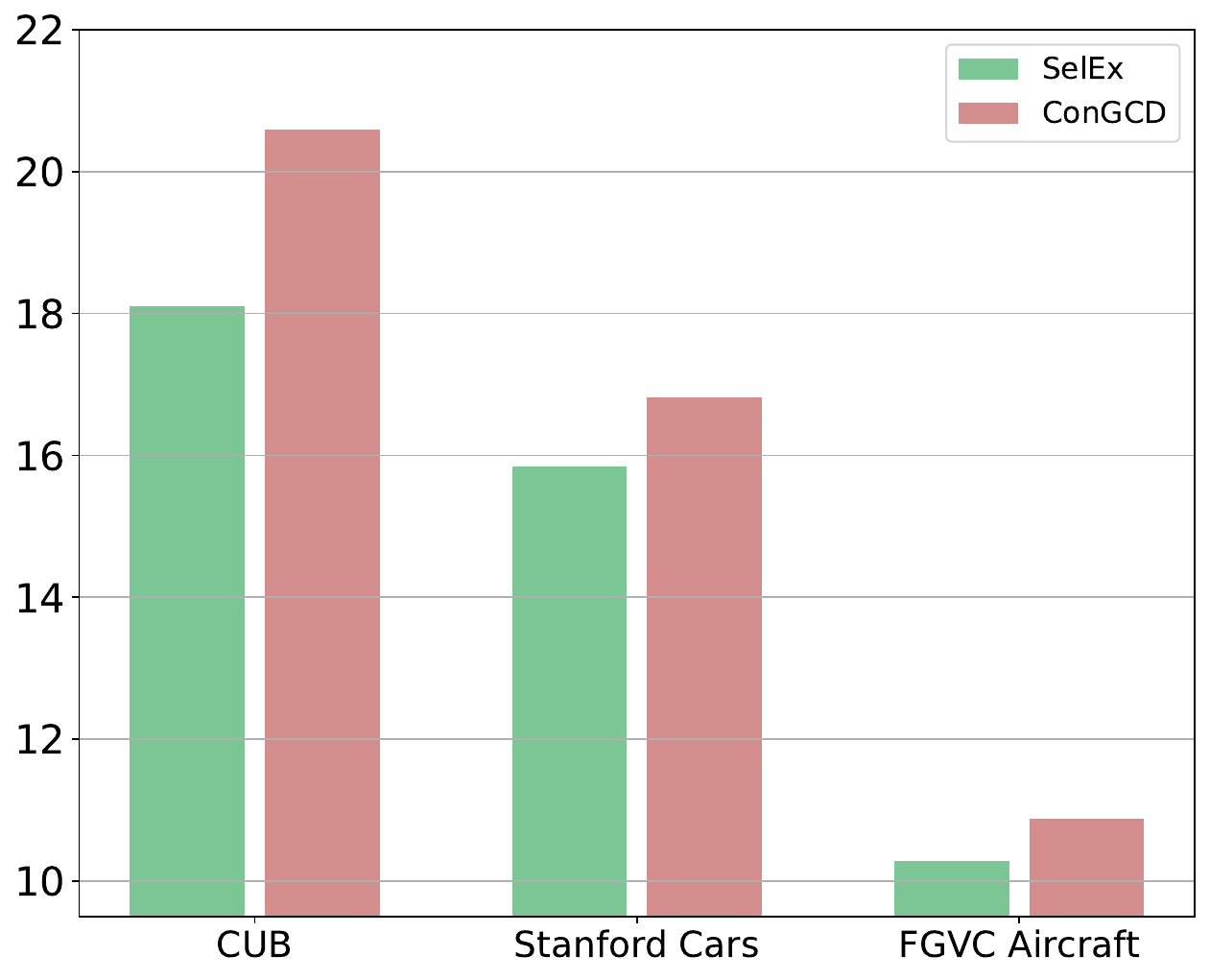}\label{fig:figure_von_HA_fine}}
  \quad
    \subfloat[$log(\operatorname{rank}(\mathcal{A}))$ of coarse-grained datasets]
  {\includegraphics[width=0.22\textwidth]{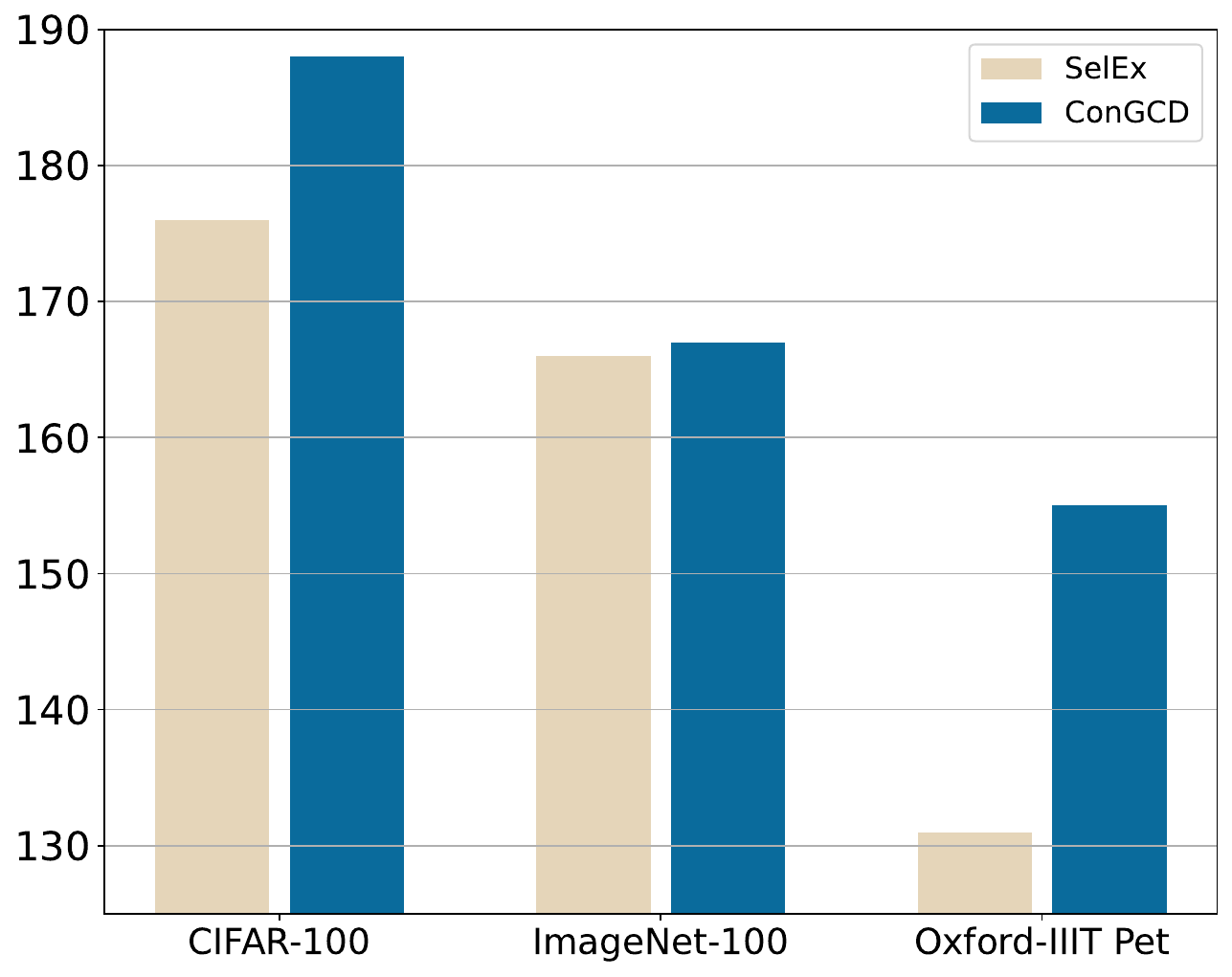}\label{fig:figure_von_log_corse}}
  \quad
    \subfloat[$\hat{H}(\mathcal{A})$ of coarse-grained datasets]
  {\includegraphics[width=0.22\textwidth]{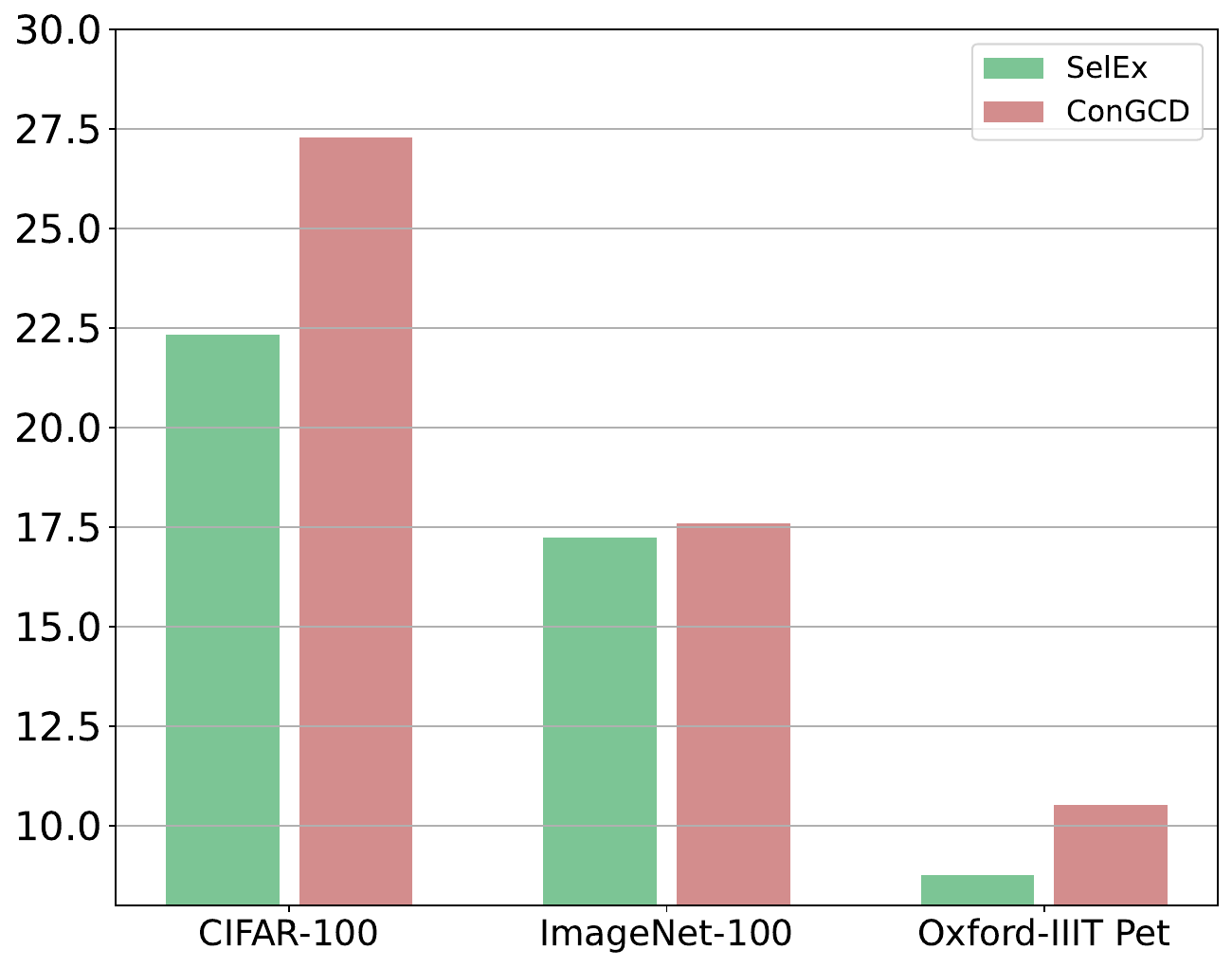}\label{fig:figure_von_HA_corse}}
  \quad
  \caption{Comparison between (a) $log(\operatorname{rank}(\mathcal{A}))$ and (b) $\hat{H}(\mathcal{A})$. The count of the largest eigenvalues necessary to account for 99\% of the total eigenvalue energy serves as a surrogate for the rank.}
  \label{fig:entropy}
\end{figure*}

\subsection{Results}
\label{subsec:results}

\noindent
\textbf{Quantitative Results.} 
ConGCD demonstrates significant improvements in both fine-grained and coarse-grained classification tasks, particularly excelling in classifying new categories while maintaining performance on known categories. In fine-grained tasks (Tab~\ref{tab:fine}), ConGCD enhances overall accuracy by an average of 1.18\% across four baselines, with accuracy on new categories increasing by 1.98\%. For coarse-grained tasks (Tab~\ref{tab:coarse}), ConGCD improves accuracy on new categories by 0.98\% on CIFAR-100, while accuracy on known categories remains stable or shows slight improvement. These results highlight ConGCD's ability to effectively classify new categories without compromising performance on known ones, demonstrating its effectiveness as a plug-and-play method with strong compatibility across diverse approaches.

\begin{figure}[h]
  \centering
  \subfloat[CUB-200]
  {\includegraphics[width=0.22\textwidth]{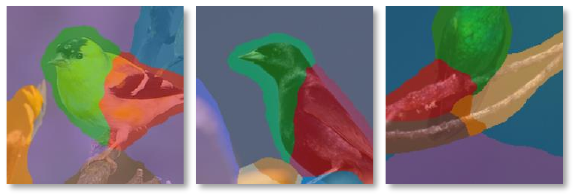}\label{fig:slotmask1}}
  \quad     
  \subfloat[FGVC-Aircraft]
  {\includegraphics[width=0.22\textwidth]{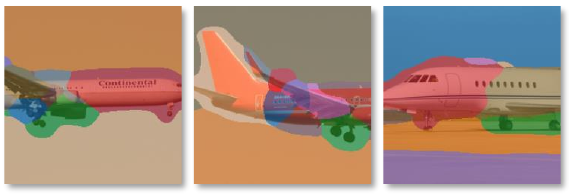}\label{fig:slotmask2}}
  \quad
  \caption{Visualization of visual primitives. In a dataset, primitives are bound to specific semantics in an unsupervised manner.}
  \label{fig:slotmask12}
\end{figure}

\begin{figure}[t]
  \centering
     \includegraphics[width=0.40\textwidth]{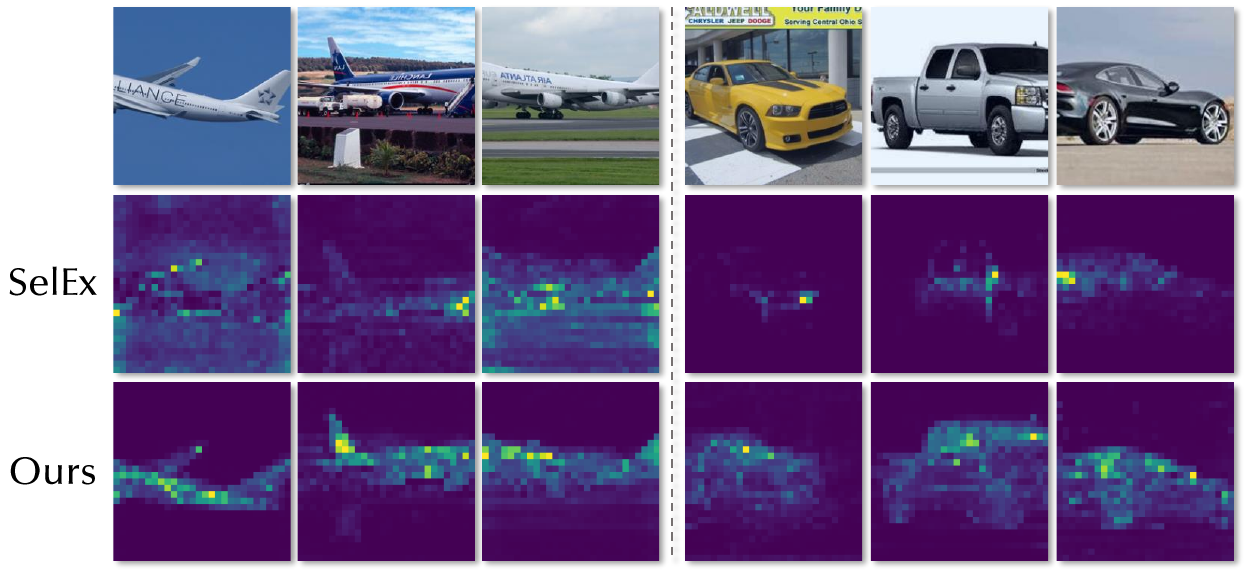}
   \caption{Visualization on attention maps of SoTA and ConGCD.}
   \label{fig:atten}
\end{figure}

\noindent
\textbf{Qualitative Results.} 
(1) Contextual consensus units capture inherent distributions, while dominant consensus units focus on class-related variables. As shown in~\cref{fig:gradcam}, weak neurons activate in low-information areas like the background, whereas dominant neurons target high-semantic difference areas (\textit{e.g.}, wheels, headlights) to enhance GCD. (2) To highlight ConGCD's performance improvement, attention maps in~\cref{fig:atten} reveal that ConGCD emphasizes the main body of the object (foreground), unlike SelEx, which focuses more on the background and specific minor areas. This enables a more uniform extraction of task-related variables. (3) As shown in~\cref{fig:slotmask12}, it can be observed that within the same dataset, visual primitives possess a high degree of cross-sample consistency. A primitive focuses on a region that has similar semantics.

\begin{figure}[th]
  \centering
  \subfloat[CIFAR-10 with SelEx]
  {\includegraphics[width=0.10\textwidth]{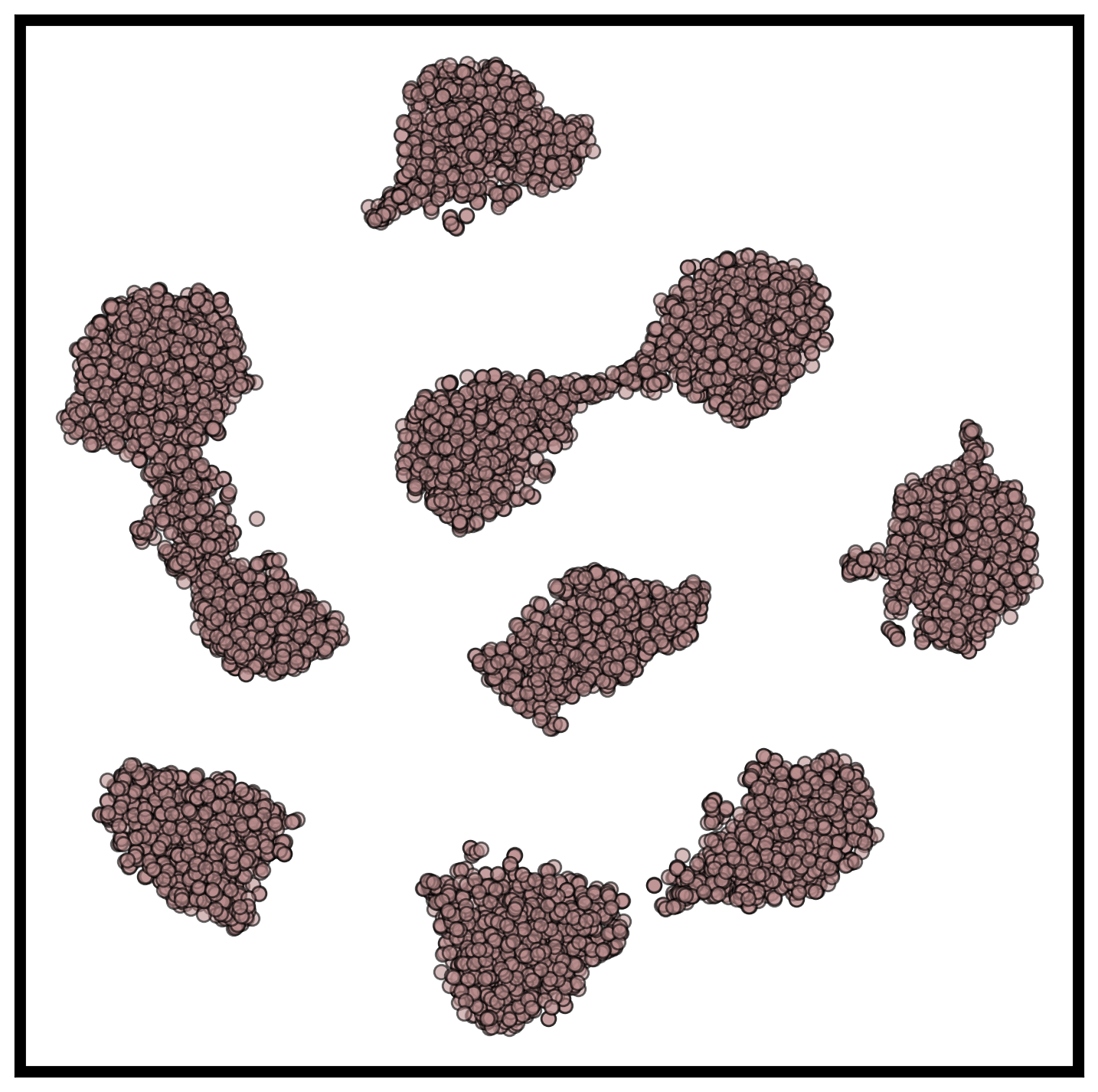}\label{fig:tsne_SelEX_cifar10}}
  \quad     
  \subfloat[CIFAR-10 with ConGCD]
  {\includegraphics[width=0.10\textwidth]{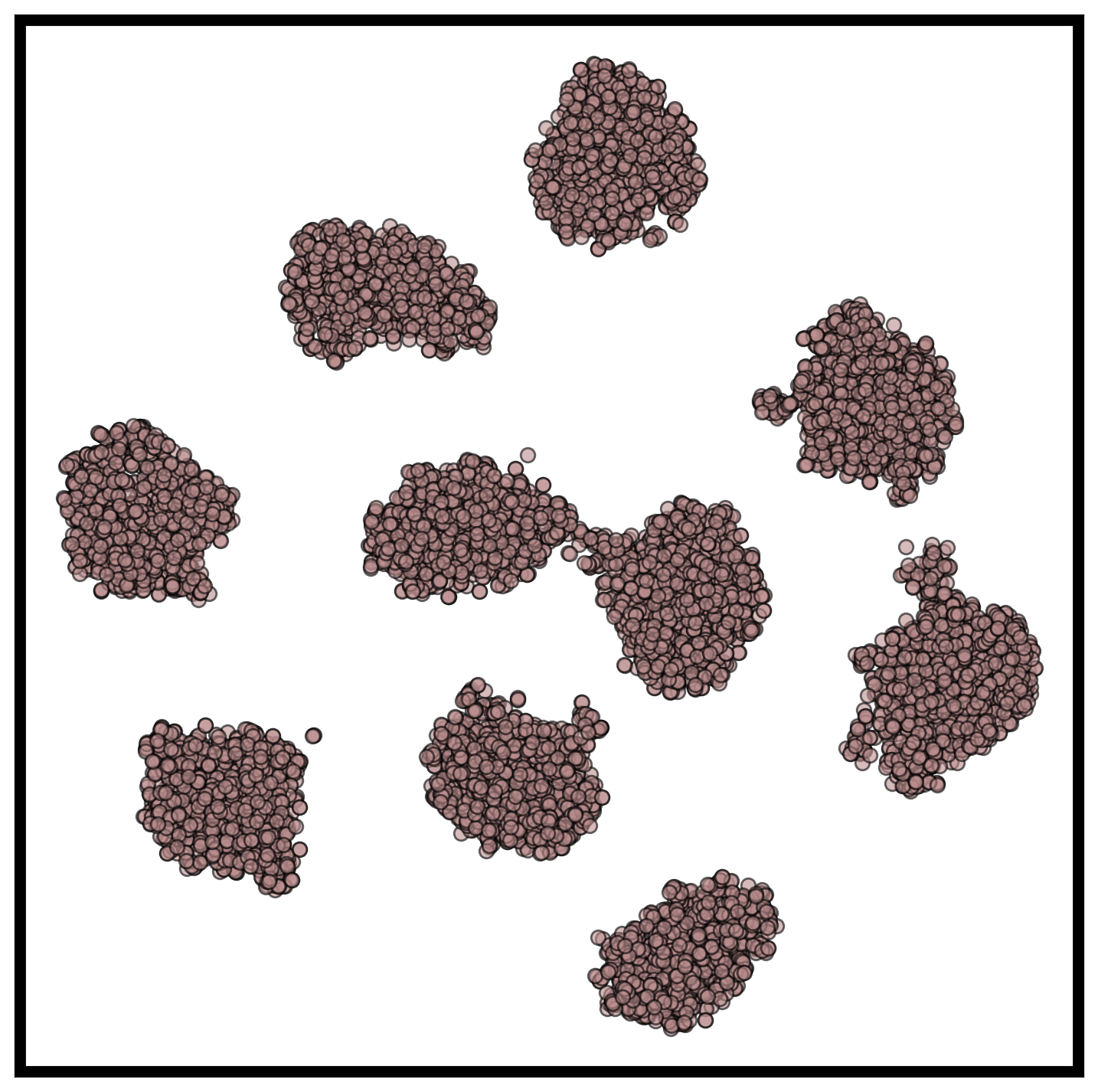}\label{fig:tsne_slotgcd_cifar10}}
  \quad
\subfloat[CIFAR-100 with SelEx]
{\includegraphics[width=0.10\textwidth]{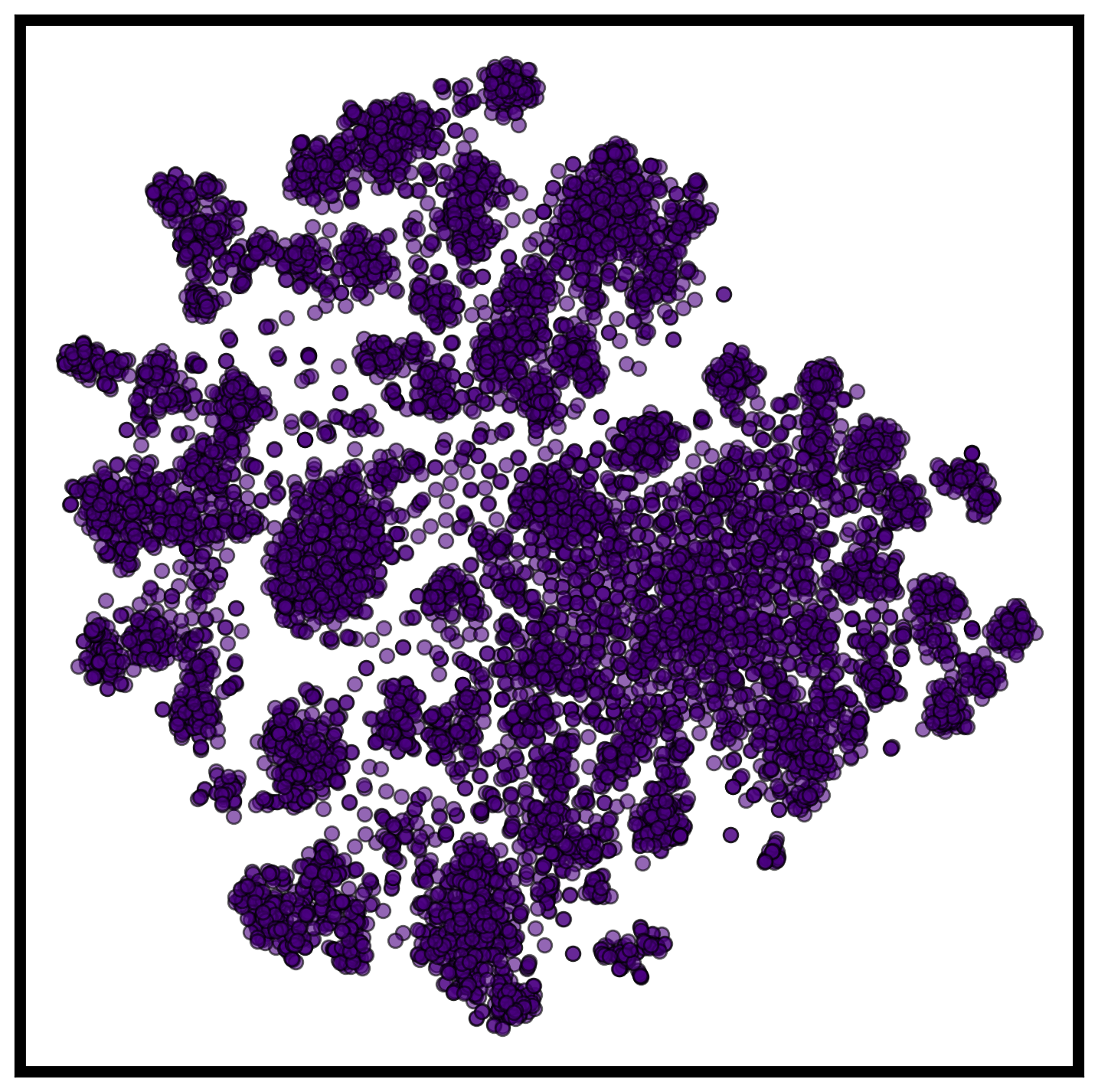}\label{fig:tsne_SelEX_cifar100}}
\quad
\subfloat[CIFAR-100 with ConGCD]
{\includegraphics[width=0.10\textwidth]{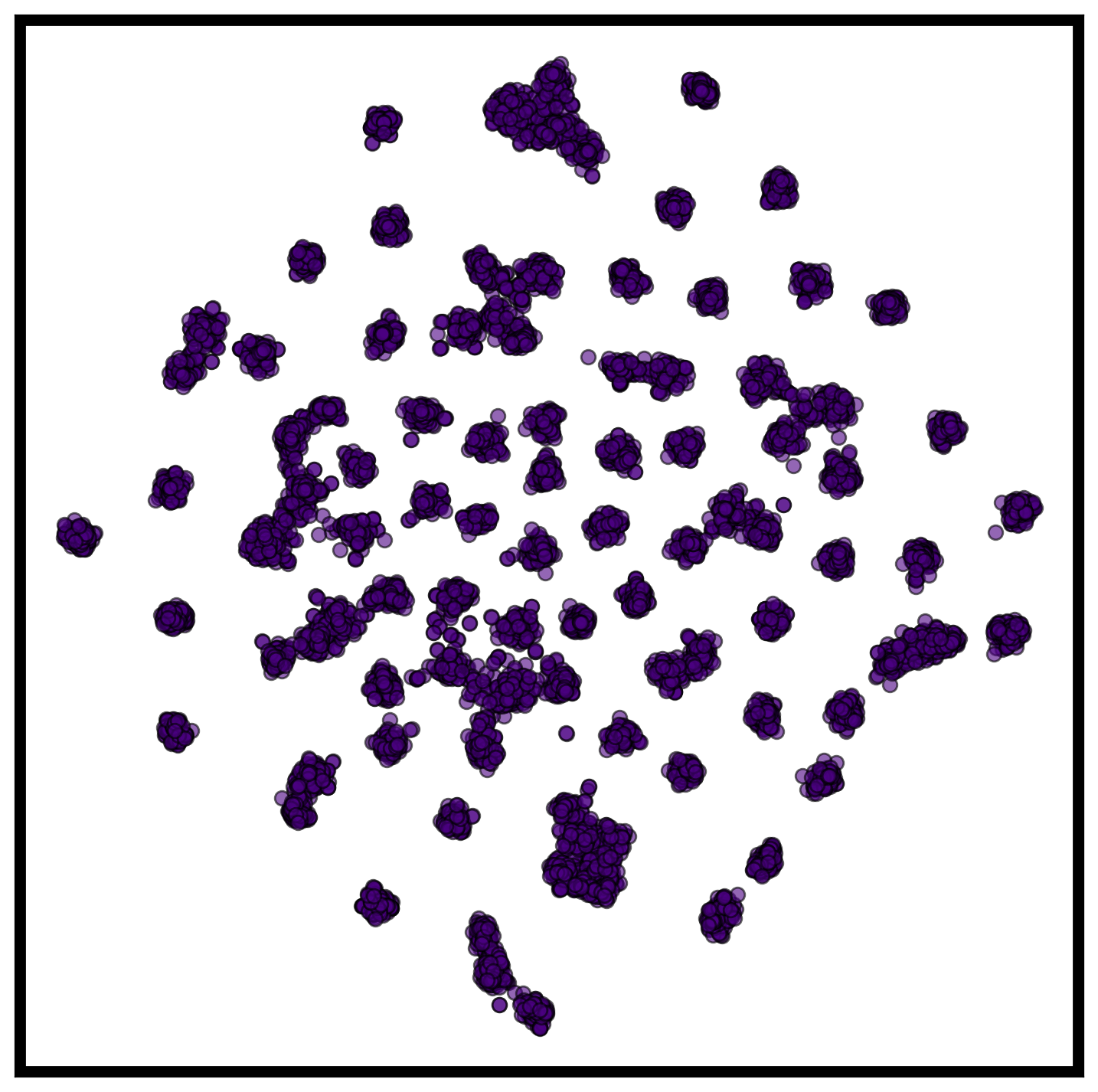}\label{fig:tsne_slotgcd_cifar100}}
\quad
  \caption{Visualisation of the embedding space with t-SNE~\cite{JMLR:v9:vandermaaten08a}.}
  \label{fig:tsne}
\end{figure}

\subsection{Analysis}
\label{subsec:analysis}

\noindent
\textbf{Visual primitives increase von Neumann entropy.} We analyze the sources of the effectiveness of ConGCD from the perspectives of representation learning and information quantification. We employ von Neumann entropy~\cite{petz2001entropy} as a measure of the amount of information in the representation. For the class token $\mathbf{z}$ used in GCD, its autocorrelation matrix $\mathcal{A} \triangleq \sum_{i=1}^N \frac{1}{N} \mathbf{z}_i \mathbf{z}_i^{\top}$. The von Neumann entropy (VNA) is expressed as $\hat{H}\left(\mathcal{A}\right) \triangleq-\sum_j \lambda_j \log \lambda_j$, where $\lambda$ is the eigenvalues of $\mathcal{A}$. As shown in~\cref{fig:entropy}, compared with SelEx, the representation of ConGCD has a higher $\hat{H}\left(\mathcal{A}\right)$ and a higher $log(\operatorname{rank}(\mathcal{A}))$, which means a larger amount of information. This indicates that ConGCD enables the GCD process to more discriminatively separate clusters by providing more discriminative features.

\noindent
\textbf{The versatility of visual primitives.} To verify the universality of ConGCD as a plug-and-play solution, we integrated ConGCD with existing GCD schemes, and the results are reported in \cref{tab:fine} and \cref{tab:coarse}. It is evident that ConGCD provides them with a consistent improvement in accuracy, which can be attributed to the unstructured decoupling of visual primitives from objects, enabling the model to possess a certain ability to model topological relationships. The model no longer solely relies on image-level embeddings from a single perspective and applies this category-independent topological relationship to the discovery of novel categories.

\begin{table}[ht]
  \centering  
  \scalebox{0.67}{
    \begin{tabular}[b]{l cc cc cc cc}
      \toprule
      \multirow{2}{*}{Method} &  
      \multicolumn{2}{c}{CIFAR-100} &
      \multicolumn{2}{c}{ImageNet-100} &
      \multicolumn{2}{c}{CUB-200} &
      \multicolumn{2}{c}{Stanford-Cars} \\
      \cmidrule(lr){2-3} \cmidrule(lr){4-5} \cmidrule(lr){6-7} \cmidrule(lr){8-9} 
      & $|\mathcal{Y}_u|$ & Err(\%) &  $|\mathcal{Y}_u|$ & Err(\%) & $|\mathcal{Y}_u|$ & Err(\%)  & $|\mathcal{Y}_u|$ & Err(\%)\\
      \midrule
      Ground truth & 100 & - & 100 & - & 200 & - & 196 & - \\
      \midrule
      GCD~\cite{vaze2022generalized} & 100 & 0 & 109 & 9  & 231 & 15.5 & 230 & 17.3\\
      DCCL~\cite{pu2023dynamic} & 146 & 46 & 129 & 29& 172 & 9 & 192 & 0.02\\  
      PIM~\cite{chiaroni2023parametric} & 95 & 5 & 102 & 2 & 227 & 13.5 & 169 & 13.8\\  
      GPC~\cite{zhao2023learning} & 100 & 0 & 103 & 3 & 212 & 6 & 201 & 0.03\\  
      \midrule
      CMS~\cite{choi2024contrastive} & 94 & 6 & 98 & 2 & 176 & 12 & 149 & 23.9 \\
      ConGCD & 97 & 3 & 99 & 1 & 186 & 7 & 160 & 18.3\\  
      \bottomrule
    \end{tabular}
  }
  \caption{Estimated number and error rate of $|\mathcal{Y}_u|$.\label{tab:kestimation}}  
\end{table}

\begin{figure*}[th]
  \centering
  \subfloat[CUB-200]
  {\includegraphics[width=0.22\textwidth]{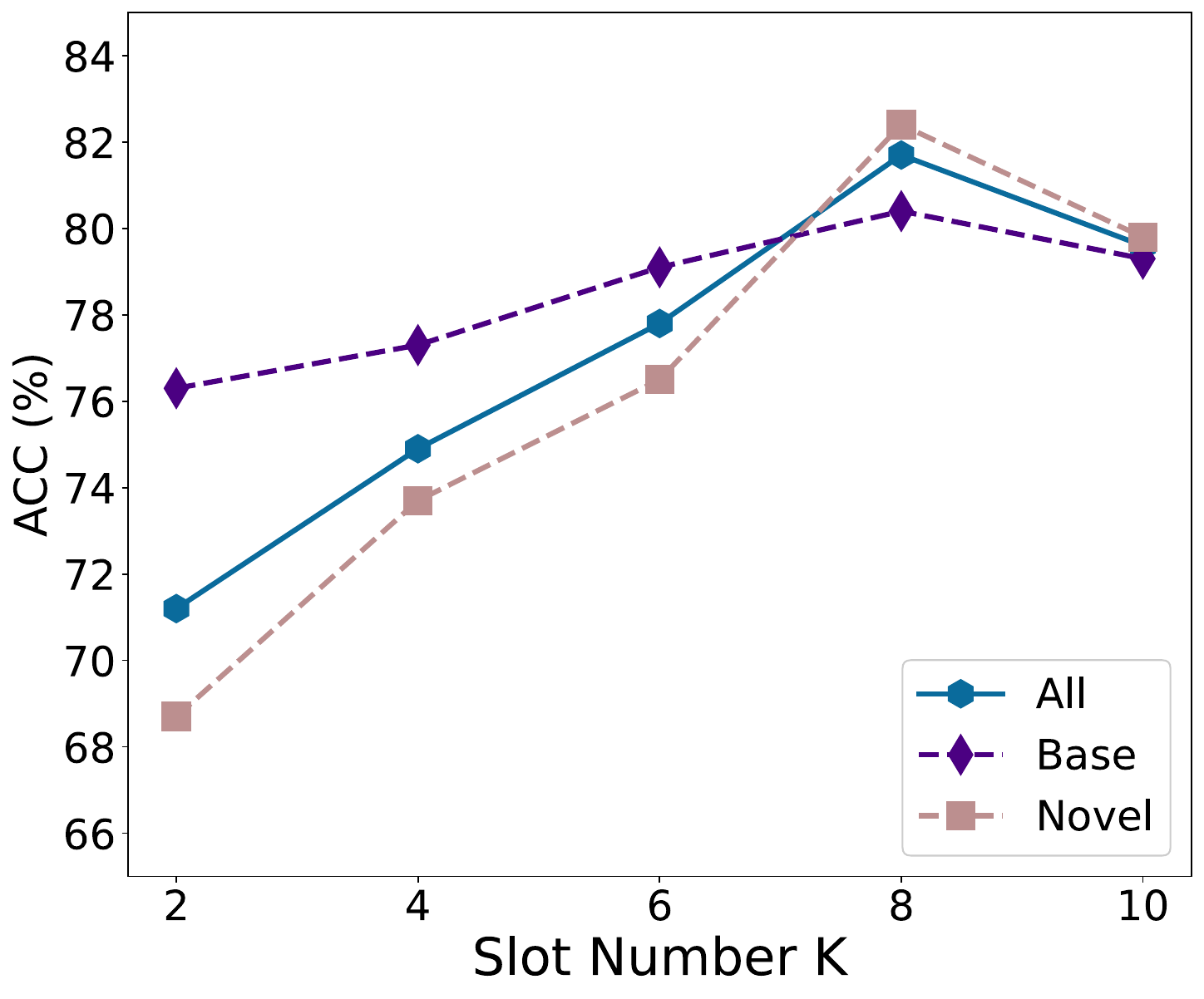}\label{fig:ablation_k_cub}}
  \quad     
  \subfloat[FGVC-Aircraft]
  {\includegraphics[width=0.22\textwidth]{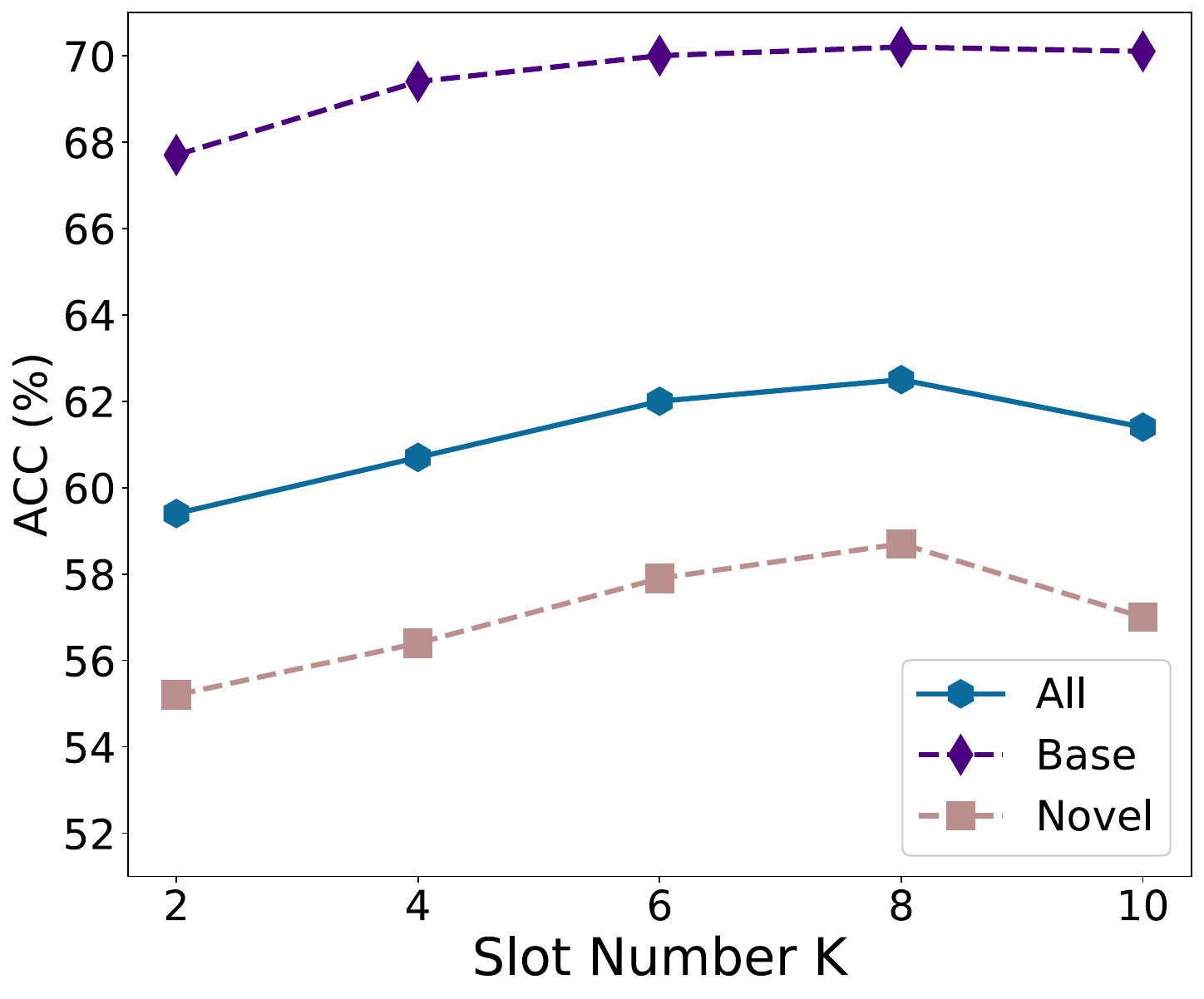}\label{fig:ablation_k_air}}
  \quad
\subfloat[CIFAR-10]
{\includegraphics[width=0.22\textwidth]{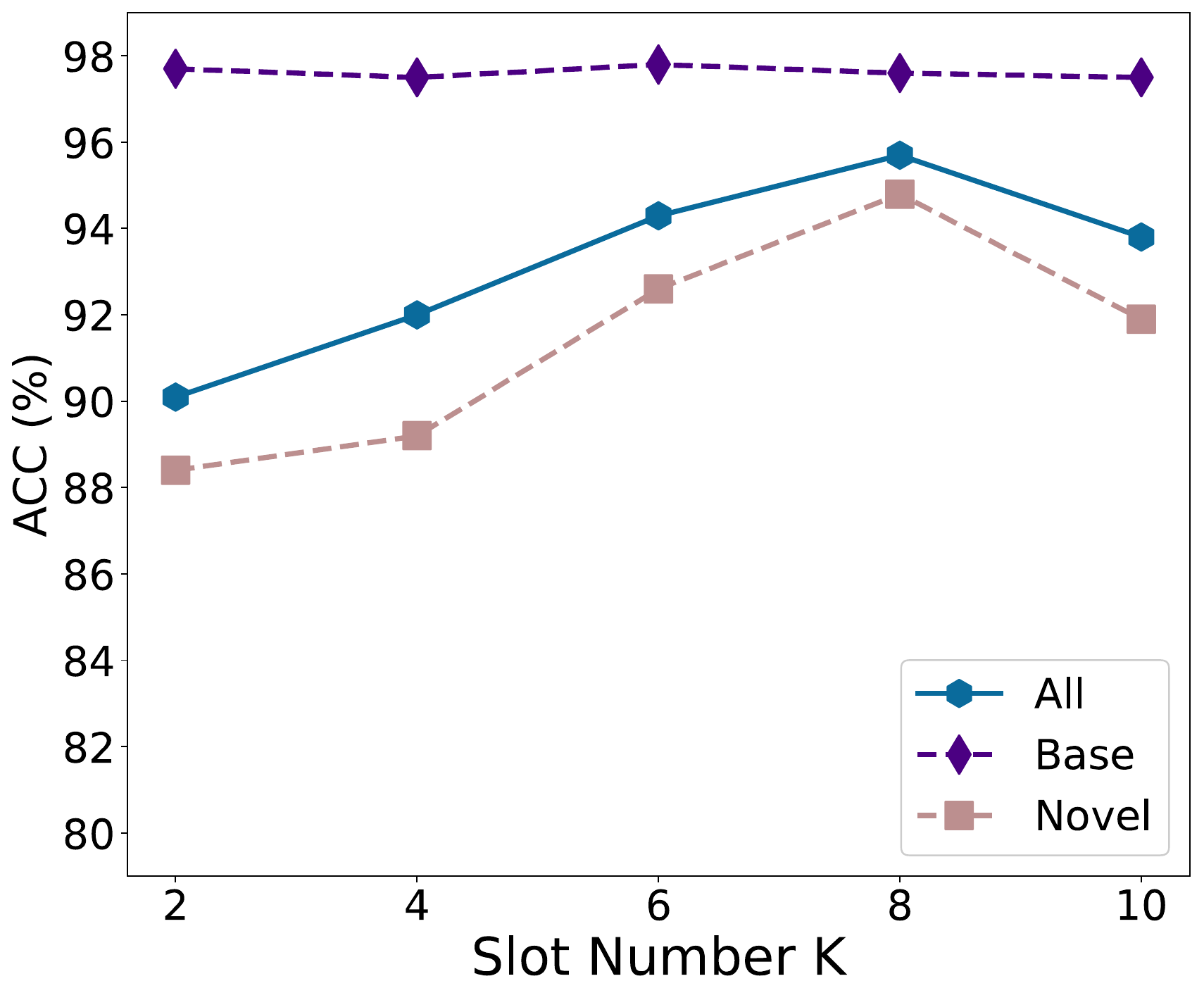}\label{fig:ablation_k_cifar10}}
\quad
\subfloat[ImageNet-100]
{\includegraphics[width=0.22\textwidth]{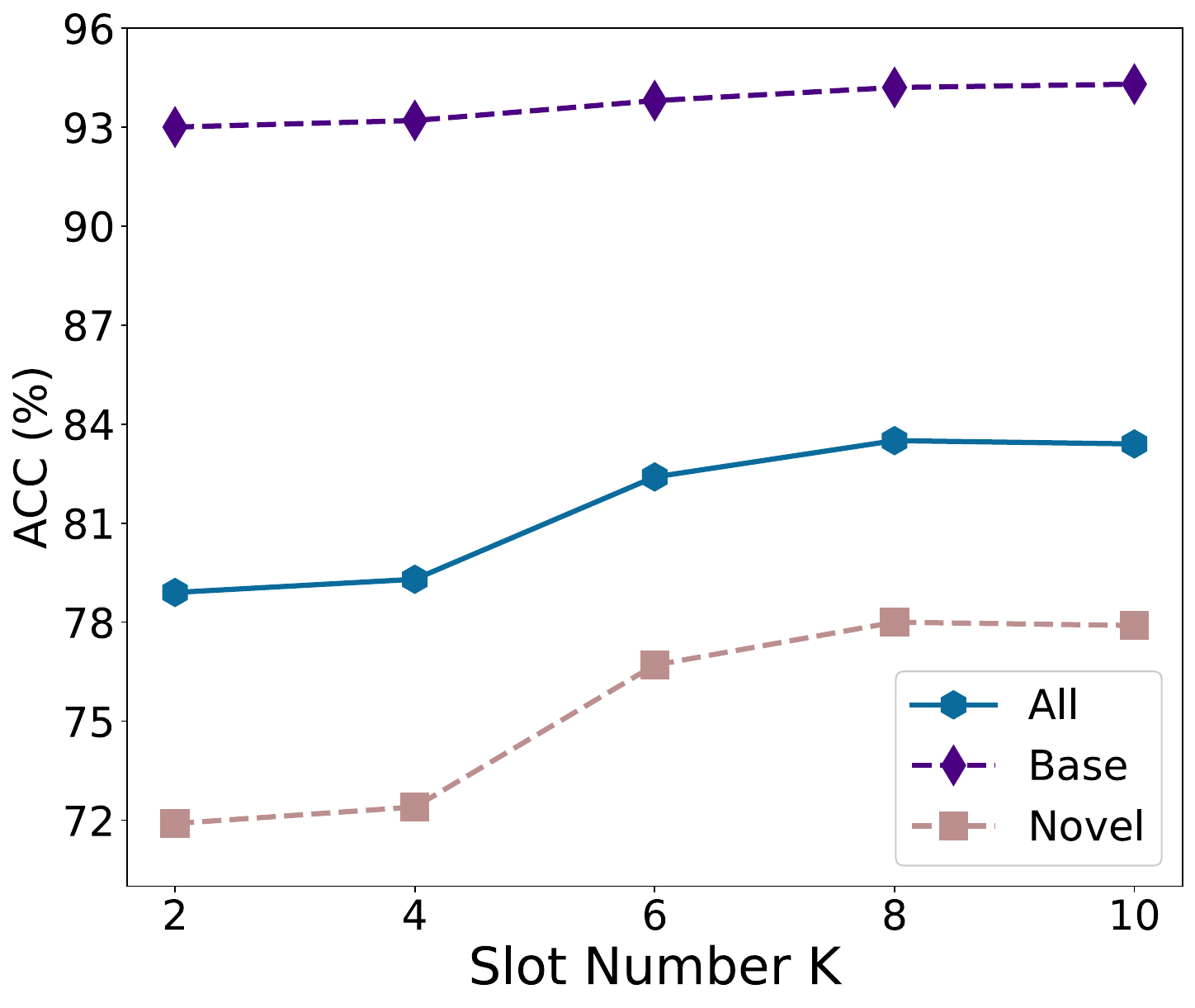}\label{fig:ablation_k_imagenet100}}
\quad
  \caption{The ablations of the number of visual primitives $K$ on (a, b) fine- and (c, d) coarse-grained benchmarks.}
  \label{fig:ablation_k}
\end{figure*}

\noindent
\textbf{ConGCD provides accurate distribution estimation.} The core of GCD is to estimate category distributions as accurately as possible and separate different categories. We evaluate ConGCD’s contribution to distribution estimation from two perspectives: clustering visualization and category number estimation. (1) We visualize embeddings using t-SNE~\cite{JMLR:v9:vandermaaten08a}, as shown in~\cref{fig:tsne}. Notably, ConGCD significantly enhances clustering quality without modifying the optimization objective. It reduces within-class distances, clarifying cluster boundaries. (2) As shown in~\cref{tab:kestimation}, ConGCD combined with CMS~\cite{choi2024contrastive} achieves the most accurate category number estimation $|\mathcal{Y}_u|$ with the lowest error rate, demonstrating its superior ability to model data and partition the feature space, crucial for reaching GCD’s upper bound.

\begin{table}[h]
  \centering
       \scalebox{0.38}
 {
\begin{tabular}{l|ccc|ccc|ccc|ccc}
\toprule
\multirow{2}{*}{\textbf{Components}}&\multicolumn{3}{c}{\textbf{CUB-200}}& \multicolumn{3}{c}{\textbf{FGVC-Aircraft}}&\multicolumn{3}{c}{\textbf{CIFAR-10}}&\multicolumn{3}{c}{\textbf{ImageNet-100}}\\ \cmidrule(lr){2-4} \cmidrule(lr){5-7} \cmidrule(lr){8-10} \cmidrule(lr){11-13}& All&Known&Novel&All&Known&Novel&All&Known&Novel&All& Known&Novel\\
\midrule
conventional MoE &78.2&79.1&77.3&59.4&78.1&53.6&93.6&97.1&91.8&80.4&92.5&74.9\\
\midrule
Ours w/o Dominant Consensus Units &81.1&80.3&81.5&62.1&69.7&58.3&94.9&97.8&94.1&82.9&94.0&77.2\\
Ours w/o Contextual Consensus Units&80.4&79.7&81.0&61.3&70.6&57.1&94.2& 97.6&92.3&81.5&93.5&75.4\\
Ours w/o Consensus Scheduler&79.8&79.5&80.0&60.9&70.1&56.7&94.4&97.7&92.7&81.4 &92.7&77.0\\

\bottomrule
\end{tabular}
}
  \caption{Ablations on components of multiplex consensus.}
  \label{tab:ablation}
\end{table} 

\noindent
\textbf{Ablations on the number of visual primitives.} As a plug-and-play solution, ConGCD requires no hyperparameter tuning, which is an important prerequisite for its wide application. The only parameter that may potentially affect the performance is the number $K$ of visual primitives. We conducted ablation studies on it for both fine- and coarse-grained datasets, as shown in~\cref{fig:ablation_k}. It can be seen that in the open world, an appropriate $K$ is sufficient to handle multiple downstream tasks. When $K$ is in the range of 7 to 10, ConGCD achieves consistent and stable performance improvements, and $K$ is set to 8 for all dataset.

\noindent
\textbf{Ablation Study on the Multiplex Consensus Mechanism.}  
We conducted an ablation study on the key component of ConGCD in~\cref{subsec:moae}, with results summarized in~\cref{tab:ablation}. The mixture of experts (MoE) is the most comparable approach to our multiplex consensus mechanism, as it directly integrates with visual primitives. However, conventional MoE proves detrimental to the GCD task, as it only accepts image-level embeddings and overfits to labeled data from known classes. The experts merely memorize features of known classes, failing to generalize to novel categories. In contrast, our multiplex consensus mechanism offers three key advantages: \textbf{(1) Dominant consensus units} process diverse visual primitives, enabling primitive-level perception and improving classification accuracy for novel categories. \textbf{(2) Contextual consensus units} capture intrinsic distributions shared across known and unknown classes, ensuring balanced clustering and mitigating bias. \textbf{(3) Consensus scheduler} introduces a learnable threshold to refine the distinction between weak and dominant neurons, optimizing unit selection and weight distribution to enhance accuracy.

\section{Conclusion}
\label{sec:con}

Current GCD approaches treat images as atomic entities, ignoring the human capacity for compositional reasoning. We propose ConGCD, a cognition-inspired paradigm bridging this gap with two key innovations: self-deconstruction decomposes objects into competing visual primitives via semantic reconstruction, while multiplex consensus dynamically integrates dual pathways, dominant consensus consensus units capturing class-specific attributes and contextual consensus units modeling distributional invariants. ConGCD achieves competitive performance on coarse- and fine-grained benchmarks, enhancing interpretability and generalization in open-world recognition.

\section*{Acknowledgement}
\label{sec:acknowledgement}
This work was supported in part by the National Natural Science Foundation of China under Grant 82172033, Grant U19B2031, Grant 61971369, Grant 52105126, Grant 82272071, and Grant 62271430; and in part by the Dreams Foundation of Jianghuai Advance Technology Center; and in part by the Open Fund of the National Key Laboratory of Infrared Detection Technologies.
{
    \small
    \bibliographystyle{ieeenat_fullname}
    \bibliography{main}

\begin{thebibliography}{60}
\providecommand{\natexlab}[1]{#1}
\providecommand{\url}[1]{\texttt{#1}}
\expandafter\ifx\csname urlstyle\endcsname\relax
  \providecommand{\doi}[1]{doi: #1}\else
  \providecommand{\doi}{doi: \begingroup \urlstyle{rm}\Url}\fi

\bibitem[Alexander(2016)]{alexander2016relational}
Patricia~A Alexander.
\newblock Relational thinking and relational reasoning: harnessing the power of patterning.
\newblock \emph{NPJ science of learning}, 1\penalty0 (1):\penalty0 1--7, 2016.

\bibitem[Aydemir et~al.(2023)Aydemir, Xie, and Guney]{aydemir2023self}
G{\"o}rkay Aydemir, Weidi Xie, and Fatma Guney.
\newblock Self-supervised object-centric learning for videos.
\newblock \emph{Advances in Neural Information Processing Systems}, 36:\penalty0 32879--32899, 2023.

\bibitem[Banerjee et~al.(2024)Banerjee, Kallooriyakath, and Biswas]{banerjee2024amend}
Anwesha Banerjee, Liyana~Sahir Kallooriyakath, and Soma Biswas.
\newblock Amend: Adaptive margin and expanded neighborhood for efficient generalized category discovery.
\newblock In \emph{Proceedings of the IEEE/CVF Winter Conference on Applications of Computer Vision}, pages 2101--2110, 2024.

\bibitem[Belloni and Robinett(2014)]{belloni2014infinite}
Mario Belloni and Richard~W Robinett.
\newblock The infinite well and dirac delta function potentials as pedagogical, mathematical and physical models in quantum mechanics.
\newblock \emph{Physics Reports}, 540\penalty0 (2):\penalty0 25--122, 2014.

\bibitem[Cao et~al.(2021)Cao, Brbic, and Leskovec]{2021Open}
Kaidi Cao, Maria Brbic, and Jure Leskovec.
\newblock Open-world semi-supervised learning.
\newblock 2021.

\bibitem[Cao et~al.(2024)Cao, Zheng, Wang, Yu, Shen, Li, Lu, and Tian]{cao2024solving}
Xinzi Cao, Xiawu Zheng, Guanhong Wang, Weijiang Yu, Yunhang Shen, Ke Li, Yutong Lu, and Yonghong Tian.
\newblock Solving the catastrophic forgetting problem in generalized category discovery.
\newblock In \emph{Proceedings of the IEEE/CVF Conference on Computer Vision and Pattern Recognition}, pages 16880--16889, 2024.

\bibitem[Caron et~al.(2021)Caron, Touvron, Misra, J{\'e}gou, Mairal, Bojanowski, and Joulin]{caron2021emerging}
Mathilde Caron, Hugo Touvron, Ishan Misra, Herv{\'e} J{\'e}gou, Julien Mairal, Piotr Bojanowski, and Armand Joulin.
\newblock Emerging properties in self-supervised vision transformers.
\newblock In \emph{Proceedings of the IEEE/CVF international conference on computer vision}, pages 9650--9660, 2021.

\bibitem[Chen(1982)]{chen1982topological}
Lin Chen.
\newblock Topological structure in visual perception.
\newblock \emph{Science}, 218\penalty0 (4573):\penalty0 699--700, 1982.

\bibitem[Chiaroni et~al.(2023)Chiaroni, Dolz, Masud, Mitiche, and Ben~Ayed]{chiaroni2023parametric}
Florent Chiaroni, Jose Dolz, Ziko~Imtiaz Masud, Amar Mitiche, and Ismail Ben~Ayed.
\newblock Parametric information maximization for generalized category discovery.
\newblock In \emph{Proceedings of the IEEE/CVF International Conference on Computer Vision}, pages 1729--1739, 2023.

\bibitem[Choi et~al.(2024)Choi, Kang, and Cho]{choi2024contrastive}
Sua Choi, Dahyun Kang, and Minsu Cho.
\newblock Contrastive mean-shift learning for generalized category discovery.
\newblock In \emph{Proceedings of the IEEE/CVF Conference on Computer Vision and Pattern Recognition}, pages 23094--23104, 2024.

\bibitem[Chung et~al.(2014)Chung, Gulcehre, Cho, and Bengio]{chung2014empirical}
Junyoung Chung, Caglar Gulcehre, KyungHyun Cho, and Yoshua Bengio.
\newblock Empirical evaluation of gated recurrent neural networks on sequence modeling.
\newblock \emph{arXiv preprint arXiv:1412.3555}, 2014.

\bibitem[Dey and Salem(2017)]{dey2017gate}
Rahul Dey and Fathi~M Salem.
\newblock Gate-variants of gated recurrent unit (gru) neural networks.
\newblock In \emph{2017 IEEE 60th international midwest symposium on circuits and systems (MWSCAS)}, pages 1597--1600. IEEE, 2017.

\bibitem[Dictionary(1989)]{dictionary1989oxford}
Oxford~English Dictionary.
\newblock Oxford english dictionary.
\newblock \emph{Simpson, Ja \& Weiner, Esc}, 3, 1989.

\bibitem[Elsayed et~al.(2022)Elsayed, Mahendran, Van~Steenkiste, Greff, Mozer, and Kipf]{elsayed2022savi++}
Gamaleldin Elsayed, Aravindh Mahendran, Sjoerd Van~Steenkiste, Klaus Greff, Michael~C Mozer, and Thomas Kipf.
\newblock Savi++: Towards end-to-end object-centric learning from real-world videos.
\newblock \emph{Advances in Neural Information Processing Systems}, 35:\penalty0 28940--28954, 2022.

\bibitem[Fei et~al.(2022)Fei, Zhao, Yang, and Zhao]{fei2022xcon}
Yixin Fei, Zhongkai Zhao, Siwei Yang, and Bingchen Zhao.
\newblock Xcon: Learning with experts for fine-grained category discovery.
\newblock \emph{arXiv preprint arXiv:2208.01898}, 2022.

\bibitem[Ferrari and Zisserman(2007)]{ferrari2007learning}
Vittorio Ferrari and Andrew Zisserman.
\newblock Learning visual attributes.
\newblock \emph{Advances in neural information processing systems}, 20, 2007.

\bibitem[Geirhos et~al.(2018)Geirhos, Rubisch, Michaelis, Bethge, Wichmann, and Brendel]{geirhos2018imagenet}
Robert Geirhos, Patricia Rubisch, Claudio Michaelis, Matthias Bethge, Felix~A Wichmann, and Wieland Brendel.
\newblock Imagenet-trained cnns are biased towards texture; increasing shape bias improves accuracy and robustness.
\newblock \emph{arXiv preprint arXiv:1811.12231}, 2018.

\bibitem[Geirhos et~al.(2019)Geirhos, Rubisch, Michaelis, Bethge, Wichmann, and Brendel]{geirhos2018imagenettrained}
Robert Geirhos, Patricia Rubisch, Claudio Michaelis, Matthias Bethge, Felix~A. Wichmann, and Wieland Brendel.
\newblock Imagenet-trained {CNN}s are biased towards texture; increasing shape bias improves accuracy and robustness.
\newblock In \emph{International Conference on Learning Representations}, 2019.

\bibitem[Geirhos et~al.(2020)Geirhos, Jacobsen, Michaelis, Zemel, Brendel, Bethge, and Wichmann]{geirhos2020shortcut}
Robert Geirhos, J{\"o}rn-Henrik Jacobsen, Claudio Michaelis, Richard Zemel, Wieland Brendel, Matthias Bethge, and Felix~A Wichmann.
\newblock Shortcut learning in deep neural networks.
\newblock \emph{Nature Machine Intelligence}, 2\penalty0 (11):\penalty0 665--673, 2020.

\bibitem[Goodman(1963)]{goodman1963statistical}
Nathaniel~R Goodman.
\newblock Statistical analysis based on a certain multivariate complex gaussian distribution (an introduction).
\newblock \emph{The Annals of mathematical statistics}, 34\penalty0 (1):\penalty0 152--177, 1963.

\bibitem[Kim et~al.(2023)Kim, Choi, Choi, and Kim]{kim2023shepherding}
Jinwoo Kim, Janghyuk Choi, Ho-Jin Choi, and Seon~Joo Kim.
\newblock Shepherding slots to objects: Towards stable and robust object-centric learning.
\newblock In \emph{Proceedings of the IEEE/CVF Conference on Computer Vision and Pattern Recognition}, pages 19198--19207, 2023.

\bibitem[Koh et~al.(2020)Koh, Nguyen, Tang, Mussmann, Pierson, Kim, and Liang]{koh2020concept}
Pang~Wei Koh, Thao Nguyen, Yew~Siang Tang, Stephen Mussmann, Emma Pierson, Been Kim, and Percy Liang.
\newblock Concept bottleneck models.
\newblock In \emph{International conference on machine learning}, pages 5338--5348. PMLR, 2020.

\bibitem[Krause et~al.(2013)Krause, Stark, Deng, and Fei-Fei]{krause20133d}
Jonathan Krause, Michael Stark, Jia Deng, and Li Fei-Fei.
\newblock 3d object representations for fine-grained categorization.
\newblock In \emph{Proceedings of the IEEE international conference on computer vision workshops}, pages 554--561, 2013.

\bibitem[Krawczyk(2012)]{krawczyk2012cognition}
Daniel~C Krawczyk.
\newblock The cognition and neuroscience of relational reasoning.
\newblock \emph{Brain research}, 1428:\penalty0 13--23, 2012.

\bibitem[Krawczyk et~al.(2011)Krawczyk, McClelland, and Donovan]{krawczyk2011hierarchy}
Daniel~C Krawczyk, M~Michelle McClelland, and Colin~M Donovan.
\newblock A hierarchy for relational reasoning in the prefrontal cortex.
\newblock \emph{Cortex}, 47\penalty0 (5):\penalty0 588--597, 2011.

\bibitem[Krizhevsky et~al.(2009)Krizhevsky, Hinton, et~al.]{krizhevsky2009learning}
Alex Krizhevsky, Geoffrey Hinton, et~al.
\newblock Learning multiple layers of features from tiny images.
\newblock 2009.

\bibitem[Krizhevsky et~al.(2012)Krizhevsky, Sutskever, and Hinton]{krizhevsky2012imagenet}
Alex Krizhevsky, Ilya Sutskever, and Geoffrey~E Hinton.
\newblock Imagenet classification with deep convolutional neural networks.
\newblock \emph{Advances in neural information processing systems}, 25, 2012.

\bibitem[Li et~al.(2024{\natexlab{a}})Li, Wu, Wang, and Han]{li2024prompt}
Deng Li, Aming Wu, Yaowei Wang, and Yahong Han.
\newblock Prompt-driven dynamic object-centric learning for single domain generalization.
\newblock In \emph{Proceedings of the IEEE/CVF Conference on Computer Vision and Pattern Recognition}, pages 17606--17615, 2024{\natexlab{a}}.

\bibitem[Li et~al.(2024{\natexlab{b}})Li, Wen, Li, and Lee]{li2024emergence}
Tianqin Li, Ziqi Wen, Yangfan Li, and Tai~Sing Lee.
\newblock Emergence of shape bias in convolutional neural networks through activation sparsity.
\newblock \emph{Advances in Neural Information Processing Systems}, 36, 2024{\natexlab{b}}.

\bibitem[Locatello et~al.(2020)Locatello, Weissenborn, Unterthiner, Mahendran, Heigold, Uszkoreit, Dosovitskiy, and Kipf]{locatello2020object}
Francesco Locatello, Dirk Weissenborn, Thomas Unterthiner, Aravindh Mahendran, Georg Heigold, Jakob Uszkoreit, Alexey Dosovitskiy, and Thomas Kipf.
\newblock Object-centric learning with slot attention.
\newblock \emph{Advances in neural information processing systems}, 33:\penalty0 11525--11538, 2020.

\bibitem[Ma et~al.(2024)Ma, Zhu, Zhong, Zhang, and Liu]{ma2024active}
Shijie Ma, Fei Zhu, Zhun Zhong, Xu-Yao Zhang, and Cheng-Lin Liu.
\newblock Active generalized category discovery.
\newblock In \emph{Proceedings of the IEEE/CVF Conference on Computer Vision and Pattern Recognition}, pages 16890--16900, 2024.

\bibitem[Maji et~al.(2013)Maji, Rahtu, Kannala, Blaschko, and Vedaldi]{maji2013fine}
Subhransu Maji, Esa Rahtu, Juho Kannala, Matthew Blaschko, and Andrea Vedaldi.
\newblock Fine-grained visual classification of aircraft.
\newblock \emph{arXiv preprint arXiv:1306.5151}, 2013.

\bibitem[Oord et~al.(2018)Oord, Li, and Vinyals]{oord2018representation}
Aaron van~den Oord, Yazhe Li, and Oriol Vinyals.
\newblock Representation learning with contrastive predictive coding.
\newblock \emph{arXiv preprint arXiv:1807.03748}, 2018.

\bibitem[Otholt et~al.(2024)Otholt, Meinel, and Yang]{otholt2024guided}
Jona Otholt, Christoph Meinel, and Haojin Yang.
\newblock Guided cluster aggregation: A hierarchical approach to generalized category discovery.
\newblock In \emph{Proceedings of the IEEE/CVF Winter Conference on Applications of Computer Vision}, pages 2618--2627, 2024.

\bibitem[Petz(2001)]{petz2001entropy}
D{\'e}nes Petz.
\newblock Entropy, von neumann and the von neumann entropy: Dedicated to the memory of alfred wehrl.
\newblock In \emph{John von Neumann and the foundations of quantum physics}, pages 83--96. Springer, 2001.

\bibitem[Pu et~al.(2023)Pu, Zhong, and Sebe]{pu2023dynamic}
Nan Pu, Zhun Zhong, and Nicu Sebe.
\newblock Dynamic conceptional contrastive learning for generalized category discovery.
\newblock In \emph{Proceedings of the IEEE/CVF conference on computer vision and pattern recognition}, pages 7579--7588, 2023.

\bibitem[Pu et~al.(2024)Pu, Li, Ji, Qin, Sebe, and Zhong]{pu2024federated}
Nan Pu, Wenjing Li, Xingyuan Ji, Yalan Qin, Nicu Sebe, and Zhun Zhong.
\newblock Federated generalized category discovery.
\newblock In \emph{Proceedings of the IEEE/CVF Conference on Computer Vision and Pattern Recognition}, pages 28741--28750, 2024.

\bibitem[Rastegar et~al.(2024{\natexlab{a}})Rastegar, Doughty, and Snoek]{rastegar2024learn}
Sarah Rastegar, Hazel Doughty, and Cees Snoek.
\newblock Learn to categorize or categorize to learn? self-coding for generalized category discovery.
\newblock \emph{Advances in Neural Information Processing Systems}, 36, 2024{\natexlab{a}}.

\bibitem[Rastegar et~al.(2024{\natexlab{b}})Rastegar, Salehi, Asano, Doughty, and Snoek]{rastegar2024selex}
Sarah Rastegar, Mohammadreza Salehi, Yuki~M Asano, Hazel Doughty, and Cees~GM Snoek.
\newblock Selex: Self-expertise in fine-grained generalized category discovery.
\newblock \emph{arXiv preprint arXiv:2408.14371}, 2024{\natexlab{b}}.

\bibitem[Ridnik et~al.(2021)Ridnik, Ben-Baruch, Noy, and Zelnik-Manor]{ridnik2021imagenet}
Tal Ridnik, Emanuel Ben-Baruch, Asaf Noy, and Lihi Zelnik-Manor.
\newblock Imagenet-21k pretraining for the masses.
\newblock \emph{arXiv preprint arXiv:2104.10972}, 2021.

\bibitem[Seitzer et~al.(2022)Seitzer, Horn, Zadaianchuk, Zietlow, Xiao, Simon-Gabriel, He, Zhang, Sch{\"o}lkopf, Brox, et~al.]{seitzer2022bridging}
Maximilian Seitzer, Max Horn, Andrii Zadaianchuk, Dominik Zietlow, Tianjun Xiao, Carl-Johann Simon-Gabriel, Tong He, Zheng Zhang, Bernhard Sch{\"o}lkopf, Thomas Brox, et~al.
\newblock Bridging the gap to real-world object-centric learning.
\newblock \emph{arXiv preprint arXiv:2209.14860}, 2022.

\bibitem[Seitzer et~al.(2023)Seitzer, Horn, Zadaianchuk, Zietlow, Xiao, Simon-Gabriel, He, Zhang, Sch{\"o}lkopf, Brox, and Locatello]{seitzer2023bridging}
Maximilian Seitzer, Max Horn, Andrii Zadaianchuk, Dominik Zietlow, Tianjun Xiao, Carl-Johann Simon-Gabriel, Tong He, Zheng Zhang, Bernhard Sch{\"o}lkopf, Thomas Brox, and Francesco Locatello.
\newblock Bridging the gap to real-world object-centric learning.
\newblock In \emph{The Eleventh International Conference on Learning Representations}, 2023.

\bibitem[Tan et~al.(2019)Tan, Liu, Ambrose, Tulig, and Belongie]{tan2019herbarium}
Kiat~Chuan Tan, Yulong Liu, Barbara Ambrose, Melissa Tulig, and Serge Belongie.
\newblock The herbarium challenge 2019 dataset.
\newblock \emph{arXiv preprint arXiv:1906.05372}, 2019.

\bibitem[Tang et~al.(2024)Tang, Yuan, Chen, Huang, Ding, and Huang]{tang2024bootstrap}
Luyao Tang, Yuxuan Yuan, Chaoqi Chen, Kunze Huang, Xinghao Ding, and Yue Huang.
\newblock Bootstrap segmentation foundation model under distribution shift via object-centric learning.
\newblock \emph{arXiv preprint arXiv:2408.16310}, 2024.

\bibitem[Tang et~al.(2025{\natexlab{a}})Tang, Huang, Chen, and Chen]{tang2025generalized}
Luyao Tang, Kunze Huang, Chaoqi Chen, and Cheng Chen.
\newblock Generalized category discovery via token manifold capacity learning.
\newblock \emph{arXiv preprint arXiv:2505.14044}, 2025{\natexlab{a}}.

\bibitem[Tang et~al.(2025{\natexlab{b}})Tang, Yuan, Chen, Zhang, Huang, and Zhang]{tang2025ocrt}
Luyao Tang, Yuxuan Yuan, Chaoqi Chen, Zeyu Zhang, Yue Huang, and Kun Zhang.
\newblock Ocrt: Boosting foundation models in the open world with object-concept-relation triad.
\newblock In \emph{Proceedings of the Computer Vision and Pattern Recognition Conference}, pages 25422--25433, 2025{\natexlab{b}}.

\bibitem[Van Den~Oord et~al.(2017)Van Den~Oord, Vinyals, et~al.]{van2017neural}
Aaron Van Den~Oord, Oriol Vinyals, et~al.
\newblock Neural discrete representation learning.
\newblock \emph{Advances in neural information processing systems}, 30, 2017.

\bibitem[van~der Maaten and Hinton(2008)]{JMLR:v9:vandermaaten08a}
Laurens van~der Maaten and Geoffrey Hinton.
\newblock Visualizing data using t-sne.
\newblock \emph{Journal of Machine Learning Research}, 9\penalty0 (86):\penalty0 2579--2605, 2008.

\bibitem[Vaswani(2017)]{vaswani2017attention}
A Vaswani.
\newblock Attention is all you need.
\newblock \emph{Advances in Neural Information Processing Systems}, 2017.

\bibitem[Vaze et~al.(2021)Vaze, Han, Vedaldi, and Zisserman]{vaze2021open}
Sagar Vaze, Kai Han, Andrea Vedaldi, and Andrew Zisserman.
\newblock Open-set recognition: A good closed-set classifier is all you need?
\newblock 2021.

\bibitem[Vaze et~al.(2022)Vaze, Han, Vedaldi, and Zisserman]{vaze2022generalized}
Sagar Vaze, Kai Han, Andrea Vedaldi, and Andrew Zisserman.
\newblock Generalized category discovery.
\newblock In \emph{Proceedings of the IEEE/CVF Conference on Computer Vision and Pattern Recognition}, pages 7492--7501, 2022.

\bibitem[Vaze et~al.(2024)Vaze, Vedaldi, and Zisserman]{vaze2024no}
Sagar Vaze, Andrea Vedaldi, and Andrew Zisserman.
\newblock No representation rules them all in category discovery.
\newblock \emph{Advances in Neural Information Processing Systems}, 36, 2024.

\bibitem[Vu et~al.(2010)Vu, Labroche, and Bouchon-Meunier]{vu2010active}
Viet-Vu Vu, Nicolas Labroche, and Bernadette Bouchon-Meunier.
\newblock Active learning for semi-supervised k-means clustering.
\newblock In \emph{2010 22nd IEEE international conference on tools with artificial intelligence}, pages 12--15. IEEE, 2010.

\bibitem[Wah et~al.(2011)Wah, Branson, Welinder, Perona, and Belongie]{wah2011caltech}
Catherine Wah, Steve Branson, Peter Welinder, Pietro Perona, and Serge Belongie.
\newblock The caltech-ucsd birds-200-2011 dataset.
\newblock 2011.

\bibitem[Wang et~al.(2024)Wang, Vaze, and Han]{wang2024sptnet}
Hongjun Wang, Sagar Vaze, and Kai Han.
\newblock Sptnet: An efficient alternative framework for generalized category discovery with spatial prompt tuning.
\newblock \emph{arXiv preprint arXiv:2403.13684}, 2024.

\bibitem[Wen et~al.(2023)Wen, Zhao, and Qi]{wen2023parametric}
Xin Wen, Bingchen Zhao, and Xiaojuan Qi.
\newblock Parametric classification for generalized category discovery: A baseline study.
\newblock In \emph{Proceedings of the IEEE/CVF International Conference on Computer Vision}, pages 16590--16600, 2023.

\bibitem[Wright(1990)]{wright1990speeding}
Mike~B Wright.
\newblock Speeding up the hungarian algorithm.
\newblock \emph{Computers \& Operations Research}, 17\penalty0 (1):\penalty0 95--96, 1990.

\bibitem[Zarlenga et~al.(2022)Zarlenga, Barbiero, Ciravegna, Marra, Giannini, Diligenti, Precioso, Melacci, Weller, Lio, et~al.]{zarlenga2022concept}
Mateo~Espinosa Zarlenga, Pietro Barbiero, Gabriele Ciravegna, Giuseppe Marra, Francesco Giannini, Michelangelo Diligenti, Frederic Precioso, Stefano Melacci, Adrian Weller, Pietro Lio, et~al.
\newblock Concept embedding models.
\newblock In \emph{NeurIPS 2022-36th Conference on Neural Information Processing Systems}, 2022.

\bibitem[Zhang et~al.(2023)Zhang, Khan, Shen, Naseer, Chen, and Khan]{zhang2023promptcal}
Sheng Zhang, Salman Khan, Zhiqiang Shen, Muzammal Naseer, Guangyi Chen, and Fahad~Shahbaz Khan.
\newblock Promptcal: Contrastive affinity learning via auxiliary prompts for generalized novel category discovery.
\newblock In \emph{Proceedings of the IEEE/CVF Conference on Computer Vision and Pattern Recognition}, pages 3479--3488, 2023.

\bibitem[Zhao et~al.(2023)Zhao, Wen, and Han]{zhao2023learning}
Bingchen Zhao, Xin Wen, and Kai Han.
\newblock Learning semi-supervised gaussian mixture models for generalized category discovery.
\newblock In \emph{Proceedings of the IEEE/CVF International Conference on Computer Vision}, pages 16623--16633, 2023.

\end{thebibliography}
}

\end{document}